%% file: ancestry_clustering.tex
\documentclass[conference]{IEEEtran}
\IEEEoverridecommandlockouts
\usepackage{cite}
\usepackage{amsmath,amssymb,amsfonts}
\usepackage{siunitx}

\usepackage{graphicx}
\usepackage{textcomp}
\usepackage{xcolor}
\usepackage[lofdepth,lotdepth]{subfig}

\usepackage{lipsum}

\def\BibTeX{{\rm B\kern-.05em{\sc i\kern-.025em b}\kern-.08em
   T\kern-.1667em\lower.7ex\hbox{E}\kern-.125emX}}

\usepackage{amssymb}
\usepackage{amsfonts}

\usepackage{anyfontsize} % allows arbitrary font sizes

\newcommand{\tinyscript}{\fontsize{6pt}{7pt}\selectfont}

\usepackage{xcolor}
\usepackage{ulem}
\usepackage[super]{nth}

\usepackage{booktabs}
\usepackage{multirow}
\usepackage{lscape}
\usepackage{makecell}
\usepackage{enumitem}
\usepackage[export]{adjustbox}

\usepackage{subcaption}

\usepackage{array}
\usepackage{booktabs}

\usepackage[colorlinks=true, urlcolor=blue, linkcolor=red]{hyperref}
\usepackage{url}

\usepackage [boxruled, rightnl]{algorithm2e}

\usepackage{algpseudocode}

\input{commands}

\DeclareMathOperator\W{W}

\begin{document}

\title{Ancestry Tree Clustering for Particle Filter Diversity Maintenance
% {\footnotesize 
% % \textsuperscript{*}Note: Sub-titles are not captured in Xplore and should not be used
% }
\thanks{\copyright 2025 IEEE. Personal use of this material is permitted. Permission from IEEE must be obtained for all other uses, in any current or future media, including reprinting/republishing this material for advertising or promotional purposes, creating new collective works, for resale or redistribution to servers or lists, or reuse of any copyrighted component of this work in other works. 

Accepted for publication at the \nth{15} International Conference on Indoor Positioning and Indoor Navigation (IPIN 2025). This online version has minor typographical and clarity fixes, and includes appendixes.}
}

\author{\IEEEauthorblockN{1\textsuperscript{st} Ilari Vallivaara}
\IEEEauthorblockA{\textit{Visiting Research Fellow} \\
\textit{University of Edinburgh}\\
Edinburgh, UK 
}
\and
\IEEEauthorblockN{2\textsuperscript{nd} Bingnan Duan}
\IEEEauthorblockA{\textit{School of Engineering} \\
\textit{University of Edinburgh}\\
Edinburgh, UK 
}
\and
\IEEEauthorblockN{3\textsuperscript{rd} Yinhuan Dong}
\IEEEauthorblockA{\textit{School of Engineering} \\
\textit{University of Edinburgh}\\
Edinburgh, UK
}
\and
\IEEEauthorblockN{4\textsuperscript{th} Tughrul Arslan}
\IEEEauthorblockA{\textit{School of Engineering} \\
\textit{University of Edinburgh}\\
Edinburgh, UK 
}
}

\maketitle

\begin{abstract}
We propose a method for linear-time diversity maintenance in particle filtering. It clusters particles based on ancestry tree topology: closely related particles in sufficiently large subtrees are grouped together. The main idea is that the tree structure implicitly encodes similarity without the need for spatial or other domain-specific metrics. This approach, when combined with intra-cluster fitness sharing and the protection of particles not included in a cluster, effectively prevents premature convergence in multimodal environments while maintaining estimate compactness. We validate our approach in a multimodal robotics simulation and a real-world multimodal indoor environment. We compare the performance to several diversity maintenance algorithms from the literature, including Deterministic Resampling and Particle Gaussian Mixtures. Our algorithm achieves high success rates with little to no negative effect on compactness, showing particular robustness to different domains and challenging initial conditions.
\end{abstract}


%% main text
\section{Introduction} \label{section:introduction}

Particle filters (PF) recursively approximate distributions over a state space using discrete weighted samples \cite{kootstra2009tackling, li2014fight}. These samples are propagated using system motion dynamics and weighted based on measurements. Resampling typically follows to produce uniform weights and concentrate computation on high-likelihood areas. Although PFs can, in principle, represent distributions of any shape, their stochastic, sample-based nature often makes it difficult to maintain the correct density over time \cite{kootstra2009tackling, li2014fight}. Loss of diversity is referred to as \textit{particle degeneracy} when it occurs in the weights, and \textit{particle impoverishment} when it occurs in the states. Diversity can be lost on a micro (local) or macro (global) level. Local loss can often be mitigated by local refinement \cite{li2014fight}. More severe global loss means no particles remain near the true state (mode). This situation is known as \textit{premature convergence}, making recovery difficult without (partial) re-initialization. Maintaining diversity is vital in multimodal environments. 

\begin{figure}[ht!]
\centering
\includegraphics[width=0.95\columnwidth]{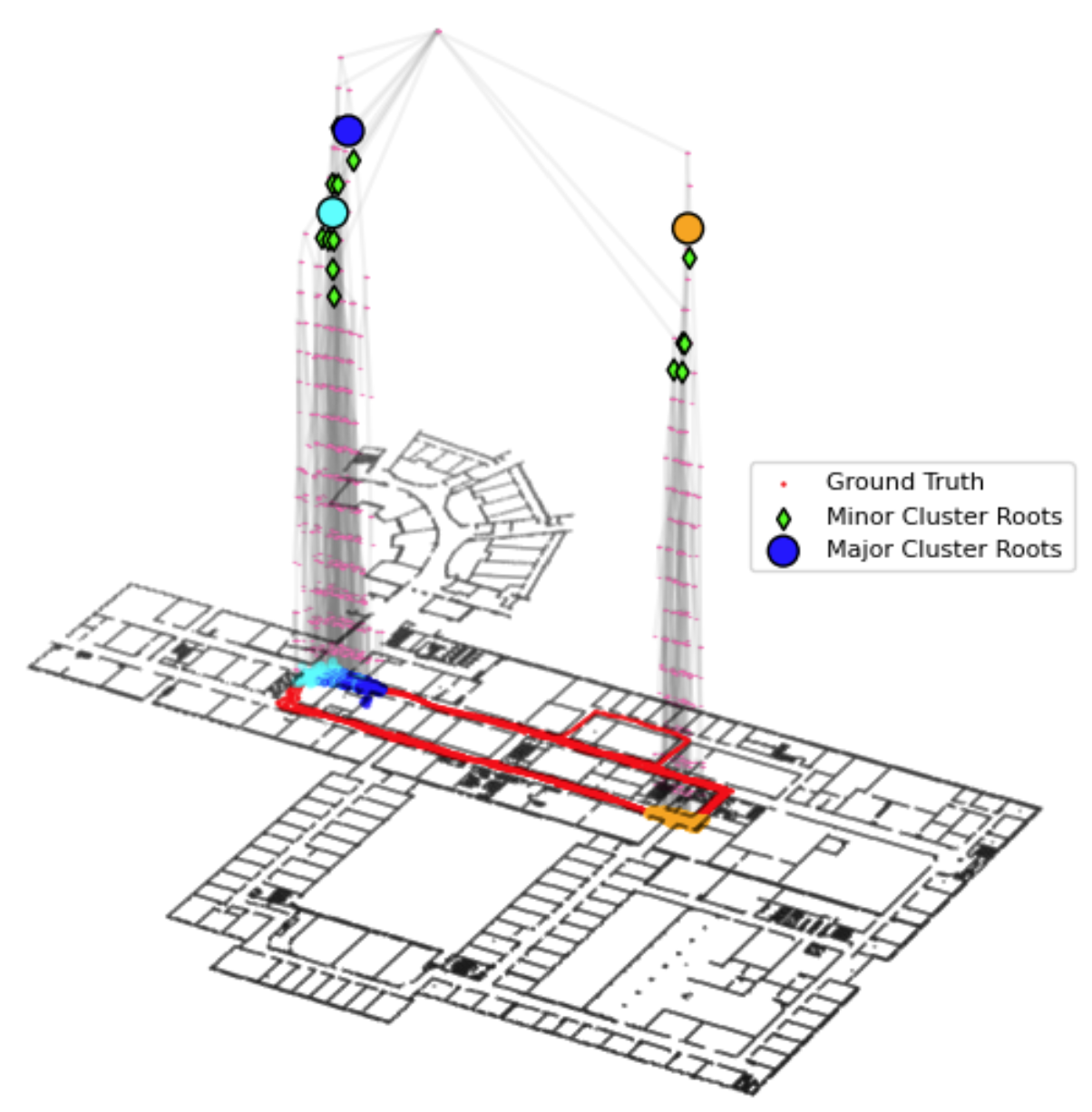}
\caption{Ancestry Tree structure capturing a multimodal distribution over a campus building floor plan during an indoor positioning experiment. The particles along the corridor are color-coded based on the subtree.}
\label{fig:ancestry_over_map}
\end{figure}

Diversity maintenance is a well-studied topic in particle filtering \cite{kootstra2009tackling, li2014fight, milstein2002robust, li2012deterministic, raihan2016particle, raihan2018particle, aguiar2021particle} and in genetic algorithms (GA) \cite{goldberg1987genetic, gabor2017genealogical, stanley2002neat_evolving}. While the simplest methods operate on weights \cite{hol2006resampling}, many consider spatial distribution via domain-specific metrics or density heuristics \cite{kootstra2009tackling, milstein2002robust, li2012deterministic, raihan2018particle, goldberg1987genetic, lindsten2011clustering}. Domain-independent approaches include estimating the genealogical distance for diversity maintenance \cite{stanley2002neat_evolving, gabor2017genealogical}.

This paper addresses maintaining explicitly multimodal distributions (macro level diversity), common in indoor positioning and robotics \cite{mendoza2019meta, davidson2010application, kaiser2011human, zampella2015indoor}. Our work explores a domain-independent approach to defining similarity and maintaining diversity. Instead of measuring spatial distance or gene differences, we tap directly into the ancestry tree topology and use the subtree structure to define similar (related) particles. We verify our approach in two simulated scenarios and one real scenario, and compare it to four other algorithms. We publish the simulation data \cite{robotics_data_link}. For a high-level view of the subtrees and their roots, Figure \ref{fig:ancestry_over_map} shows a snapshot of the ancestry tree topology in a real-world positioning experiment.

The main contributions of this paper are:
\begin{enumerate}[label=(\alph*)]
    \item We introduce a simple, computationally cheap method to extract clusters of approximate size $k$ from the ancestry tree topology. It has
    \begin{enumerate}[label=(\roman*), leftmargin=*, labelwidth=2em, align=left]
        \item Linear-time complexity in the particle count
        \item A very straightforward implementation (specifically the simplest variant)
    \end{enumerate}
    \item We show that, when applied to multimodal positioning, this clustering has several desirable properties:
    \begin{enumerate}[label=(\roman*), leftmargin=*, labelwidth=2em, align=left]
        \item Effective and adaptive diversity maintenance
        \item Little loss in compactness as a trade-off
        \item Robustness to initial conditions
        \item Robustness to environmental variation
    \end{enumerate}
\end{enumerate}

Our intuition is that, despite its simplicity, our domain-independent approach captures similarity in a fundamentally novel way, distinct from domain-specific metrics. It generalizes across domains without parameter tuning and lays a solid foundation for future refinements.

\section{Background}
\subsection{Particle Filters and Resampling}
Particle filters (PF) recursively approximate distributions over a state space $S$ using $P$ discrete weighted samples $\{p^{(i)}_t\}_{i=1}^P$, where $p^{(i)}_t = (\xIT, \wIT) \in S \times \mathbb{R}$ \cite{kootstra2009tackling, li2014fight}. The \textit{motion model} $p(\xIT \mid \xITPrev, \u_t)$ predicts state evolution given the control input $\u_t$, and the \textit{measurement model} $p(\z_t \mid \x_t)$ evaluates consistency with the current observation $\z_t$. The target posterior $p(\x_t \mid \z_{1:t})$ is approximated by the empirical distribution $\sum_{i=1}^P w_t^{(i)} \delta(\x_t - \xIT)$, where $\delta(\cdot)$ is the Dirac delta function.  To avoid particle degeneracy, where only a few particles have contributing weight, updates are typically followed by a resampling step that creates a new particle set with uniform weights. Common algorithms include multinomial, residual, and systematic resampling \cite{hol2006resampling}. This step is often triggered when the estimated effective sample size $N_{\text{eff}}$ falls below a threshold.

\subsection{Ancestry Trees}
Any resampling method RS, producing the next set of particles $\{p^{(i)}_t\}_{i=1}^P \xrightarrow{\text{RS}} \{p^{(i)}_{t+1}\}_{i=1}^P$, assigns each particle zero or more offspring. Recording these child relations over time $t = 1, \dots, T$ forms an ancestry tree, where each node represents a particle and its children are its direct offspring. If, at $t=0$, we initialize the particles as children of a dummy root, we end up with a single tree instead of a forest. The tree evolves stochastically with the filter, and the current population is represented by alive leaves. Unmaintained, the tree will grow in $O(PT)$. However, we can maintain the tree by removing dead branches (no alive leaves) and recursively merging nodes with only one child. When kept minimal in this manner, it grows and shrinks dynamically, with a maximum size of $2P$ \cite{eliazar2004dp}. Maintaining a minimal tree has a linear cost in $P$ \cite{eliazar2005hierarchical}.

Ancestry trees have been applied to particle filtering for various purposes. They provide compact map representations for Rao-Blackwellized PF, both for grid \cite{eliazar2004dp, eliazar2005hierarchical} and scattered data maps \cite{vallivaara2018quadtree}, and efficiently store paths \cite{jacob2015path}. They also enable smoothing through ancestor relations \cite{lindsten2012ancestor}. A key advantage is \textit{coalescence} \cite{jacob2015path, eliazar2004dp}, where the population's common histories are combined and distant history eventually collapses into a single root (\textit{trunk}), while the evolving subtrees (\textit{crown}) store recent history. Under certain assumptions, this allows storing the entire path history in $O(T + P \log P)$ memory, compared to the naive $O(TP)$ \cite{jacob2015path}. This is supported by empirical evidence \cite{eliazar2004dp, vallivaara2018quadtree}. 

\subsection{Diversity maintenance for PFs}
Maintaining diversity involves a trade-off between exploration and exploitation, often conflicting with unimodal and multimodal performance \cite{kootstra2009tackling, milstein2002robust}. \textit{Particle Density Optimization} (PDO) aims to improve filter density through various strategies, including local refinement, intelligent resampling strategies \cite{kootstra2009tackling, li2012deterministic}, clustering \cite{raihan2018particle, lindsten2011clustering}, fitness sharing \cite{kootstra2009tackling}, and Gaussian Mixture Models \cite{raihan2018particle, psiaki2016gaussian}. We refer to \cite{li2014fight} for a review.

Beyond weighting, density can be adjusted by moving or resampling particles. Local refinement, such as \textit{jittering/roughening}, \textit{MCMC moves}, and gradient-based methods \cite{li2014fight, boopathy2024resampling}, aim to improve the density on a local level, while low-variance resampling (e.g., systematic resampling) maintains diversity at the weight level \cite{hol2006resampling}. More advanced algorithms also consider the spatial distribution of particles \cite{li2012deterministic, raihan2018particle, lindsten2011clustering}.

State-based (spatial) \textit{clustering} is a popular approach in (global) PF diversity maintenance \cite{milstein2002robust, lindsten2011clustering, raihan2018particle, aguiar2021particle}. It is related to \textit{niching} in GAs which protects subpopulations from selection pressure \cite{goldberg1987genetic, gabor2017genealogical, stanley2002neat_evolving}, and there is a crossover between the disciplines \cite{kootstra2009tackling, li2014fight}. If you can reliably identify clusters (modes) and their count, you can explicitly maintain low-likelihood clusters.  However, finding the optimal clustering and domain-specific similarity is far from trivial.  PDO often comes with a significant computational cost and domain-specific parameter tuning \cite{li2014fight, kootstra2009tackling, raihan2016particle}. Although spatial closeness is commonly used, computing domain-specific similarity between samples can be challenging or ill-defined. Gabor et al. \cite{gabor2017genealogical} propose a domain-independent alternative for GAs by measuring genealogical distance between genomes augmented with "trash genes" that do not contribute to fitness but aid in tracking diversity.

\section{Algorithm - ATOG}
Our intuition is that support over histories corresponds to support over the state space, and that coalescence is linked to sample impoverishment: each time we prune the tree, diversity is lost. For example, in the trunk, diversity is reduced to one sample. This is a central idea of this paper: if we can detect and control coalescence, we may use it as a tool to maintain diversity. Specifically, we use the (local) ancestry tree topology to define clusters (niches). These clusters can be used in fitness-sharing-like algorithms. We name our algorithm \textit{Ancestry tree ToploloGy clustering}, or shortly \textit{ATOG}.

\subsection{Clusters based on tree topology}

For our ancestry tree, we define $c(n)$ as the children of node $n$. A \textit{descendant} of node $n$ is a node $m$ for which we can find a chain of child relations $n \xrightarrow{c(\cdot)} \dots \xrightarrow{c(\cdot)} m$. Inversely, if that is the case, $n$ is an \textit{ancestor} of $m$.
We define \textit{subtree weight} for each node recursively as 
\begin{align} \label{eq:tree_weighs}
\W(n) = \begin{cases} 
  1, & \text{if $n$ is a leaf} \\
  \sum\limits_{m \in \text{c(n)}} \W(m), & \text{if $n$ is a branch}
  \end{cases}
\end{align}
We note that this particular weighting corresponds to the subtree leaf count. Now, we want to define clusters of approximately size $k$. For this, we use a simple rule to find nodes that we call \textit{cluster roots} $\{C(k)^R_j\}$, where $j = 1, \dots, n_R$. Node $n$ is a \textit{cluster root} if 
\begin{equation} \label{eq:tree_clusters}
\begin{aligned}
    &\text{(a)} \quad \W(n) \geq k, \text{ and} \\
    &\text{(b)} \quad \forall m \in c(n) : \W(m) < k
\end{aligned}
\end{equation}
In other words, the cluster roots are the subtree roots after which the weight drops below $k$ for the first time. We note that $n_R \leq P/k$. Similarly, we define \textit{cluster leaves} $C(k)_j$ for cluster $j$ as all leaves that are descendants of  $C(k)^R_j$. Since leaves represent particles, the terms are used interchangeably. The exact cluster size depends on branching factors; however, in all our experiments, the sizes consistently fall between $\left[k, 1.5k\right]$ (Fig. \ref{fig:appendix_tree_clusters}).

\begin{figure}[t]
\centering
\subfloat[][Tree topology]{
\includegraphics[width=0.45\columnwidth]{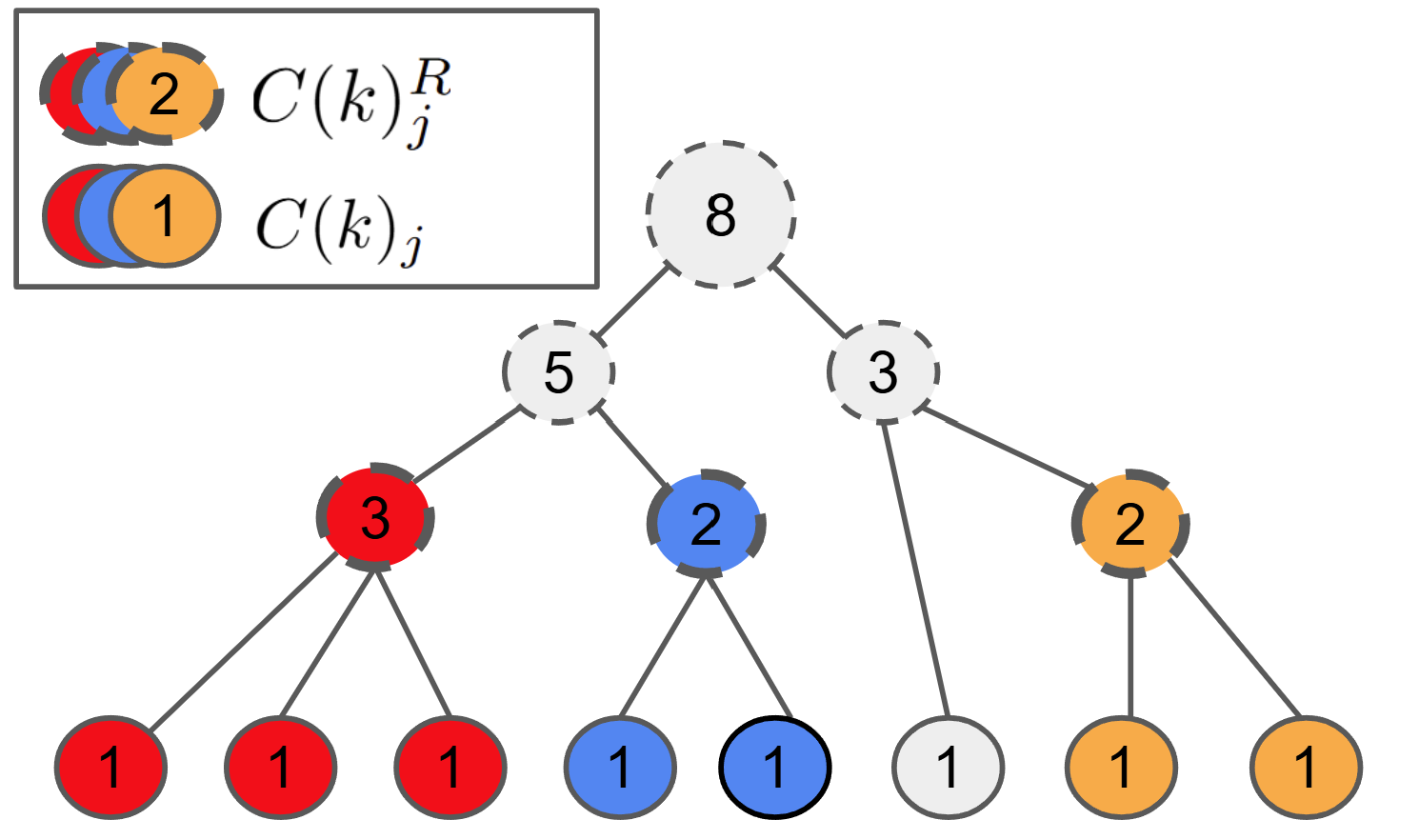}
\label{fig:ancestry_tree_pos}}
\subfloat[][Tree as path history]{
\includegraphics[width=0.45\columnwidth]{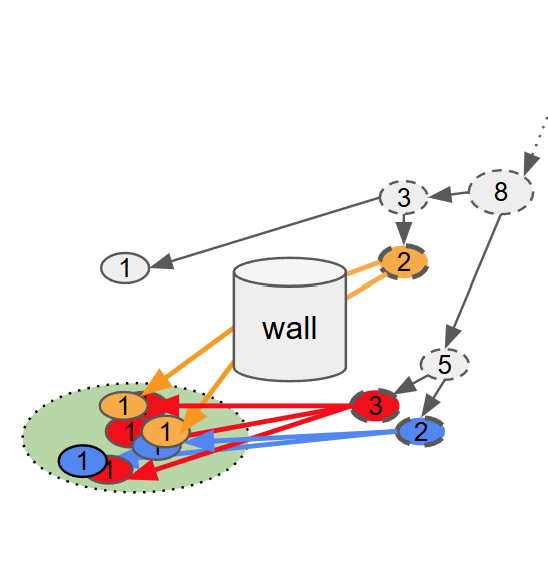}
\label{fig:ancestry_tree_struct}}
\caption{Conceptual visualization of an ancestry tree after three resampling steps, depicting cluster roots and color-matched leaves. Numbers indicate the accumulated leaves (weight) in each subtree. Here, the cluster threshold is $k = 2$. Spatial clustering (green) would not capture path histories.}
\label{fig:ancestry_tree}
\end{figure}

Figure \ref{fig:ancestry_tree} illustrates this cluster definition for an example ancestry tree. As the ancestry tree has at most $2P$ nodes \cite{eliazar2004dp}, all of these steps can be computed and updated in $O(P)$. Finally, any particle $p \notin \bigcup_{j=1}^{n_R} \{C(k)_j\}$ is assigned to a special cluster $C(k)_0$, which is of particular interest for diversity maintenance later. For brevity, we will write $C_j := C_j(k)$ if $k$ is not relevant to the context.

\subsection{Cluster-Dependent Selection -- protecting the non-cluster particles in $C_0$}
A subtree with many nodes indicates (local) convergence. To maintain diversity, we want to reward particles in smaller subtrees. This is vital in the early stages of positioning, as demonstrated later. To promote diversity, we use a reward function that we call \textit{Cluster-Dependent Selection} (CDS): 

\begin{align}
w^{(i)}_{\text{CDS}} = \begin{cases} 
  \lambda_0 \cdot w^{(i)}_t, & \text{if $p^{(i)} \in \ C_0$} \\
  1 \cdot w^{(i)}_t, & \text{otherwise.}
  \end{cases} \label{eq:cluster_dependent_selection}
\end{align}
Here, $\lambda_0 > 1$ is a parameter balancing diversity and compactness. We note that $\lambda_0$ could be any function ($> 1$), but for simplicity we define it as a constant. We optimize its value in Section \ref{sec:sensitivity}.

\subsection{Intracluster fitness sharing} \label{section:fitness_sharing}
Fitness sharing within niches or clusters can be implemented in various ways, from particle replacement strategies to more complex approaches 
\cite{goldberg1987genetic, milstein2002robust, kootstra2009tackling}. We implement the following weight update strategy for particles belonging to cluster $C_j$: 

\begin{align}
w^{(i)}_t &= p(\z_t | \x_t^{(i)})  w^{(i)}_{t-1} \label{eq:likelihood} \\
% w^{(i)}_{C_j} &= P\left[|C_j|\sum_{c \in C_j} w^{(c)}_t\right]^{(-1)} w^{(i)}_t \\
w^{(i)}_{C_j} &= \frac{|C_j| w^{(i)}_t}{P \sum_{c \in C_j} w^{(c)}_t} \label{eq:normalize}
\end{align}

First, we compute the measurement likelihood $p(\z_t | \x_t^{(i)})$ (\ref{eq:likelihood}). Then, we normalize weights so that each cluster's total weight corresponds to its relative size $|C_j|/P$ (\ref{eq:normalize}). Essentially, particles with a matching cluster index (including $0$) share fitness with each other. For this, we utilize previously developed machinery for conditional likelihood comparison \cite{vallivaara2013monty}. Other ATOG-based fitness sharing strategies are expected to perform similarly.

\subsection{Population-wide selection pressure ($C_0$ and tax)}\label{section:tax}
Note that we include $0$ in the fitness sharing indices, so that non-cluster particles in $C_0$ do get initial selection pressure, too. This pressure effectively prevents \textit{ghost particles}, while $C_0$ itself is protected by the weight adjustment (\ref{eq:cluster_dependent_selection}). In addition, we form an additional \textit{Inheritance Tax} cluster $C_{\text{tax}}$, which randomly includes particles with $0.05$ probability (the tax rate). This gradually shares fitness from clusters to the entire population and is crucial to eventually starve clusters stuck in a dead end (evolutionary or literal).

\subsection{Major and Minor clusters}
In an ideal world, an environment with $s$ symmetric locations should produce $s$ clusters. For a balanced tree, the cluster size threshold could be set as $k \approx 0.9 \cdot P/s$. However, specifically early in the run, the ancestry tree may lack a clear structure. Further, particles covering one location might have completely separate histories (Fig. \ref{fig:ancestry_tree_pos}). Also, some room must be left for non-cluster particles. We found that a smaller threshold, $k = 0.05P$, gives excellent results, even without expecting $20$ ambiguous locations. The idea of overclustering is similar to the Blob Filter \cite{psiaki2016gaussian}. This leads to defining \textit{major (M)} and \textit{minor (m)} clusters, with thresholds $k_M = 0.15P$ and $k_m = 0.05 P$, respectively. Minor clusters capture local structure and convergence, while major clusters emerge later. For our experiments, we use only minor clusters, but note that it would be straightforward to use major clusters in addition to balance particles between main modes.

\section{Algorithms chosen for comparison}
We implement a total of six algorithms for our experiments: four from literature with different diversity maintenance strategies \cite{kootstra2009tackling, li2012deterministic, raihan2018particle, raihan2016particle} and two variants of the proposed ATOG approach, one with and one without fitness sharing. We only describe the algorithms at a high level and refer to the original papers for details:

\subsubsection{\textbf{The standard particle filter (PF)}} Acts as a baseline and replicates the results in \cite{kootstra2009tackling}. Like with all of our applicable algorithms, we resample if $N_{\text{eff}}$ drops below $0.95P$ and use systematic resampling \cite{hol2006resampling}.

\subsubsection{\textbf{Frequency-Dependent Selection (FDS)}}
\label{section:freq_dep_sel}
Frequency-Dependent Selection is a niching PF variant replicated from \cite{kootstra2009tackling} that re-weights particles by multiplying their weight by the distance to a random sample of $k$ elements:
\begin{align} \label{eq:frequency_dependent_selection}
\hat{w}^{(i)}_t = w^{(i)}_t \cdot \sum_{j=0}^{k} d(i, \text{rand}(0, n)),
\end{align}
where $d(i, j)$ represents the Euclidean distance between particles $p^{(i)}$ and $p^{(j)}$. We implement this algorithm for comparison due to its simplicity and diversity-maintaining properties. It has characteristic flaws in compactness, particularly in unimodal tracking. Following \cite{kootstra2009tackling}, we set $k = 0.2 P$ relative to the particle count $P$, leading to a $O(n^2)$ complexity.

\subsubsection{\textbf{Deterministic Resampling PF (DR-SIR)}} The DR-SIR algorithm  \cite{li2012deterministic} maintains diversity by protecting low-density regions through quadtree-based \cite{vallivaara2018quadtree} spatial partitioning (Fig. \ref{fig:dr-sir_and_pgm_behavior}). It combines deterministic residual resampling (\textit{copy particles}) with support cell selection based on residual weights remaining in quadtree cells. \textit{Support particles} are generated by averaging residual weights if they exceed a minimum threshold (we set $w_{\text{min}} = 0.0001$). This gives a new particle set usually close to original set in size. To keep $P$ constant, we either fill the new set with systematic resampling or select $P$ random elements without replacement. We use 2D spatial partitioning, and set the maximum cell width to $L_{\text{start}} = 2.0$, subdivision limit to $L_{\text{min}} = 0.25$, and bucket capacity to $\alpha = 8$.

\subsubsection{\textbf{Particle Gaussian Mixture II (PGM-II)}}
PGM-II \cite{raihan2018particle} uses a sample-based approach for non-linear propagation and measurement, and fits a \textit{Gaussian Mixture Model} (GMM) for resampling. At each time step: (i) propagate the prior as samples; (ii) cluster them into $M_t^*$ Gaussian mixands; (iii) draw samples from the mixands using MCMC; (iv) weight the samples by the measurement model; (v) compute weights for the mixands from the sample weights; (vi) form a weighted mixture as the posterior and draw the final samples from it. 

For pose distance, we combine the Euclidean and angle distances. For cluster fit, we use Sum-of-norms clustering \cite{lindsten2011clustering} with $\lambda=2.0$ instead of the naive strategy suggested in \cite{raihan2016particle}.  We follow their approach and brute force for best-fit count $M_t^* \in \{1, \dots, M_{\text{max}}\}$ for clusters found by k-means. This scales as $O(PM_{\text{max}}^2)$ \cite{raihan2016particle}. We start the run with $M_{\text{max}} = 20$, and slowly decay the upper bound towards $M_t^* + 2$ to manage computational load.

\subsubsection{\textbf{ATOG Fitness Sharing (ATOG-FS)}}
This is the full version of our algorithm that     
\begin{enumerate} [label=(\alph*)]
    \item Protects non-cluster particles with CDS (\ref{eq:cluster_dependent_selection})
    \item Shares fitness mainly within clusters (\ref{section:fitness_sharing})
    \item Gradually shares inter-cluster weight via taxation (\ref{section:tax})
\end{enumerate}

\subsubsection{\textbf{ATOG Cluster Dependent Selection (ATOG-CDS)}} This is a simplified version that is very close to the standard PF. The only difference is that we adjust the weights by rewarding particles not belonging to a cluster \eqref{eq:cluster_dependent_selection}.

\begin{table}[h!]
\centering
\tinyscript
\setlength{\tabcolsep}{2pt} % tighter horizontal padding
\renewcommand{\arraystretch}{1.05} % slightly compressed rows
\begin{tabular}{|p{0.12\columnwidth}|p{0.3\columnwidth}|p{0.52\columnwidth}|}
\hline
\textbf{Abr.} & \textbf{Name and info} & \textbf{Main idea} \\
\hline
\textbf{PF} & Standard Particle Filter \cite{kootstra2009tackling} & Baseline that performs poorly in maintaining diversity.  \\
\hline
\textbf{FDS} & Frequency-Dependent Selection \cite{kootstra2009tackling} & Distance-based heuristic to give more weight to particles in less dense areas \eqref{eq:frequency_dependent_selection}. Uses $0.2P$ random samples. \\
\hline
\textbf{DR-SIR} & Deterministic Resampling SIR PF \cite{li2012deterministic} & Quadtree-based spatial partitioning based on particle density. Resampling selects initial particles with the deterministic step of residual resampling and then adds low-density support grid cells based on the cells' residual weights.  \\
\hline
\textbf{PGM-II} & Particle Gaussian Mixture II \cite{raihan2018particle} & Propagates prior as samples, then fits a GMM based on dynamically determined spatial clusters. Evaluates non-linear likelihoods for clusters using MCMC to obtain a weighted posterior. This is used to create samples for the next time step. \\
\hline
\textbf{ATOG-FS} & Ancestry Tree Topology Clustering Fitness Sharing (proposed) & Clusters particles based on ancestry tree topology. \textit{(i)} Mostly intra-cluster fitness sharing \textit{(ii)} Small amount of inter-cluster fitness sharing via taxation \textit{(iii)} Heuristically gives more weight to particles with no cluster.  \\
\hline
\textbf{ATOG-CDS} & Ancestry Tree Topology Clustering Cluster-Dependent Selection (proposed) & Simplified version of ATOG-FS: uses only \textit{(iii)} to protect small clusters \\
\hline
\end{tabular}
\caption{Description of selected algorithms.}
\end{table}

\section{Experiment 1: Diversity Maintenance in Robotics Simulation}\label{section:experiment1}
Kootstra et al. \cite{kootstra2009tackling} evaluate diversity maintenance in a robotics simulator using one explicitly multimodal and one unimodal environment. It is one of the few studies that evaluates premature convergence in a controlled and replicable manner. We publish our data for even easier reproducibility \cite{robotics_data_link}. We replicate their experimental setup and implement a subset of their algorithms along with others from the literature. The square-symmetric ambiguous environment \textit{Square} prevents the robot from distinguishing between four identical locations. The desired behavior is to maintain all modes (Fig. \ref{fig:env_square}). In contrast, the non-ambiguous \textit{Maze} should yield a unimodal distribution (Fig. \ref{fig:env_maze}). To match the experiments in \cite{kootstra2009tackling}, we run each algorithm 50 times across particle counts $P \in \{200, \dots, 2500\}$ in both environments. To assess replication accuracy, we also report numbers from the original paper (Table \ref{table:kootstra_results}), including the \textit{Closest of the Worst} algorithm (Cl. o/t w.) with top multimodal performance. We do not implement it due to its poor unimodal behavior. Our PF and FDS success rates and convergence times are slightly better, but overall trends and other metrics match.

\subsection{Experimental setup}
\subsubsection{Environment and Robot}
\begin{figure}[t]
\centering
\subfloat[][Square (ambiguous)]{
\raisebox{-.5\height}{ \includegraphics[width=0.48\columnwidth] {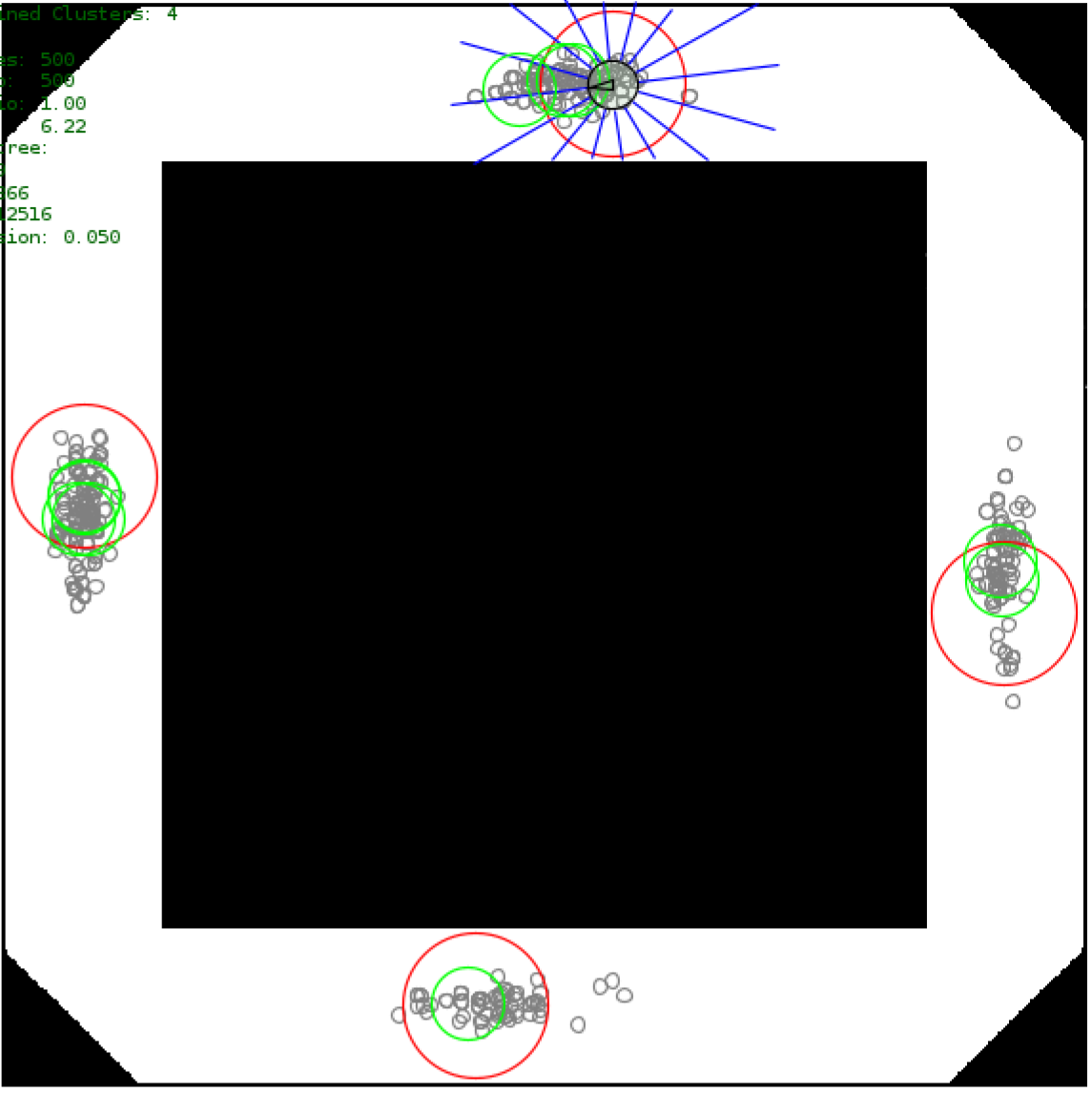}}
\label{fig:env_square}}
\subfloat[][Maze (non--ambiguous)]{
\raisebox{-.5\height} {\includegraphics[width=0.48\columnwidth]{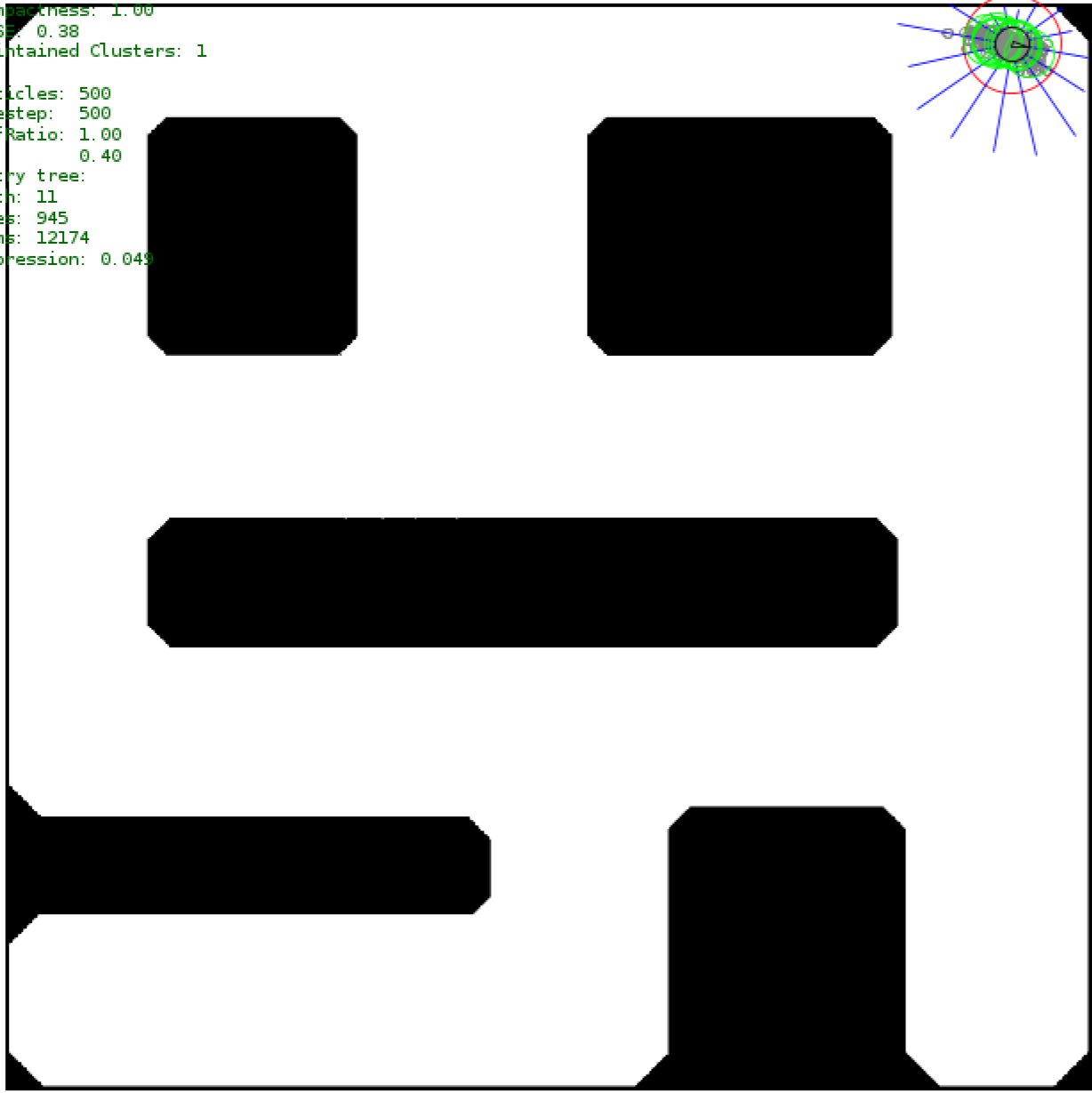}}
\label{fig:env_maze}}
\caption{The two environments used in our robotics experiments and the desired behavior. In the Square environment, we aim for the particles (grey) to maintain all four modes (red circles). In the Maze environment, we want to converge to the unambiguous right position (red circle with the robot). The green circles depict ATOG minor clusters.}
\label{fig:environments}
\end{figure}

We scale the $150 \times 150$ unit environments from \cite{kootstra2009tackling} by 0.1 for more interpretable, metric-like units compatible with algorithms assuming metric scales. The differential-drive robot has a radius of $0.35$ and a 16-ray distance sensor (range 2) with Gaussian noise ($\sigma_s = 0.1$). It moves using a Braitenberg-style controller in $0.8$ unit steps, with odometry subject to Gaussian noise ($\sigma_{trans} = 0.1$, $\sigma_{rot} = 0.04$). Figure \ref{fig:environments} illustrates the robot and the environments with the desired behavior.

\subsubsection{Initialization} Each run spans 500 time steps. The robot’s initial position between runs is randomized using a random motion sequence followed by a 0.5-probability direction flip. Particles are systematically initialized on a grid covering the drivable space, with randomized orientation. This structured initialization may explain the slightly higher success rates we observe.

\subsubsection{Motion and sensor model}
To model uncertainty, we double the noise levels used in simulation. For the sensor model, each ray is modeled with a Gaussian ($\sigma_{sm} = 2\sigma_s = 0.2$). The likelihood of particle $p^{(i)}_t$ is computed as the product of $N(z(k)_t - z(k)^{(i)}_t; 0, \sigma_{sm}^2)$ across rays, where $z(k)_t$ is the observed and $z(k)^{(i)}_t$ the simulated ray length for ray $k$.

\subsection{Goals and metrics}
\subsubsection{Diversity maintenance} 
The goal is to maintain all modes (either 4 or 1). A mode is considered non-maintained if all particles are lost in its vicinity ($d < 1.0$) for at least 50 steps. A run is considered \textit{successful} if all modes are maintained (Fig. \ref{fig:env_square}).
\subsubsection{Compactness and RMSE}
We also aim for the cloud to be dense around the modes. \textit{Compactness} is defined as  the fraction of particles within 1.0 of the closest mode. We compute RMSE over each particle's distance to the closest mode. As these are highly inversely correlated, we primarily report only RMSE. Note that low RMSE alone does not imply good performance; for example, the Standard PF achieves low RMSE by converging to a randomly selected mode.

\subsection{Sensitivity analysis of $\lambda_0$ on success rate and RMSE} \label{sec:sensitivity}
\begin{figure}[ht]
\centering
\includegraphics[width=0.95\columnwidth]{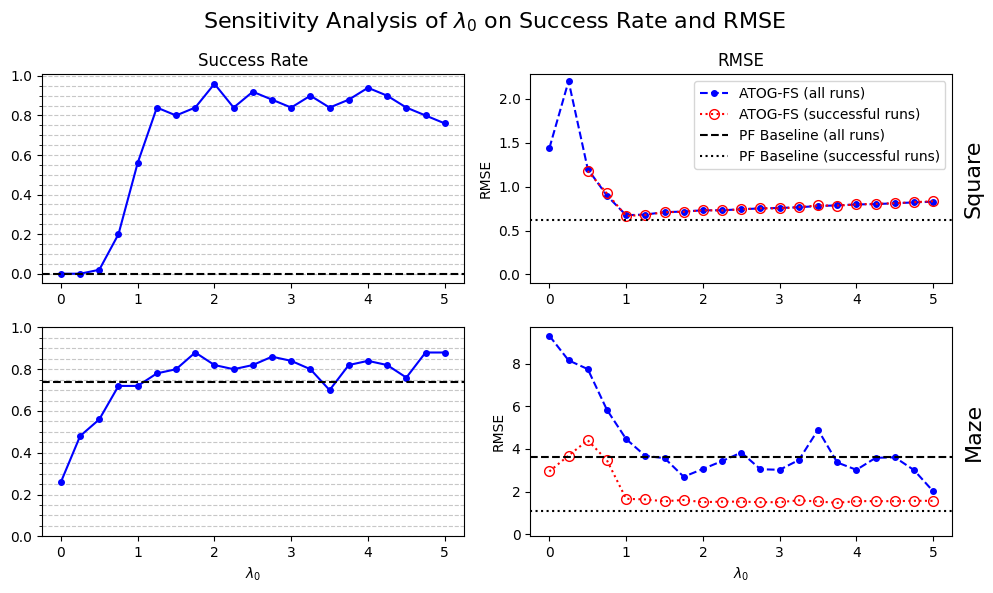}
\caption{Sensitivity analysis of $\lambda_0$ parameter that protects the non-cluster particles.   % Based on the results, we set $\lambda_0 = 2.0$ in all remaining experiments.
}
\label{fig:sensitivity}
\end{figure}

To find suitable values for $\lambda_0$, we brute-force through a quantized parameter space $\lambda_0 \in \{0.0, 0.25, \dots, 5.0 \}$. For each value, we run positioning 50 times in the Square and Maze environments using ATOG-FS with $P=1000$ and measure both success rate and RMSE. Figure \ref{fig:sensitivity} illustrates that success rates fluctuate slightly but reach peak at around $\lambda_0 = 2.0$; RMSE increases slowly but steadily after $\lambda_0 = 1.0$. The algorithm is not sensitive to exact parameter values for $\lambda_0 \in \left[2.0, 4.0\right]$, so we set $\lambda_0 = 2.0$ in all remaining experiments. 

\subsection{Results}

\subsubsection{Square environment}

\begin{figure}[ht!]
\centering
\includegraphics[width=0.95\columnwidth]{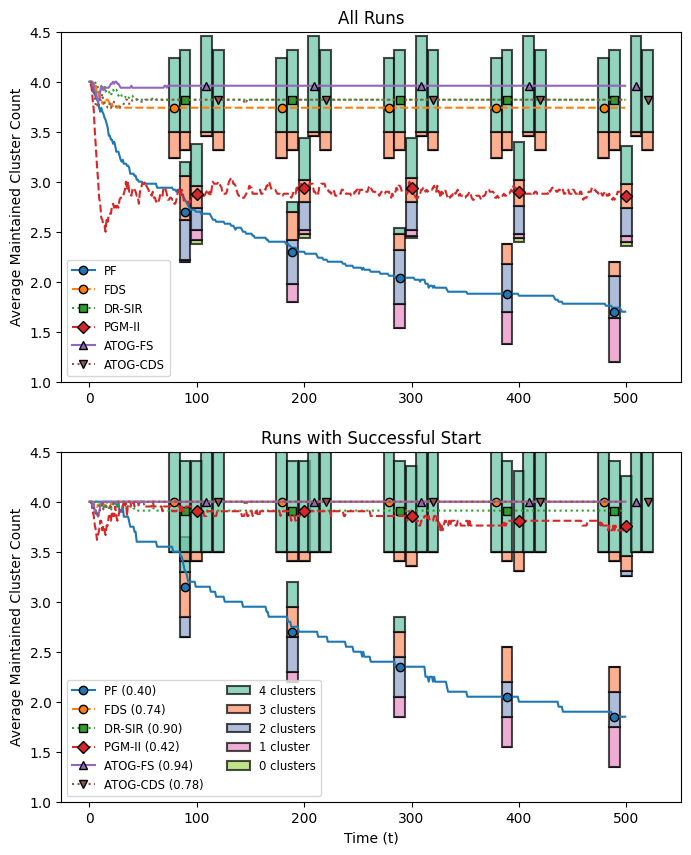}
\caption{Average maintained cluster counts in the Square environment and their distribution over time. The bottom figure legend shows the ratio of successful starts for each algorithm.}
\label{fig:maintained_clusters}
\end{figure}

\begin{figure}[ht]
    \centering
    % First row
    \subfloat[DR-SIR ($t=0$)\label{fig:dr-sir_t0}]{
        \includegraphics[width=0.47\columnwidth]{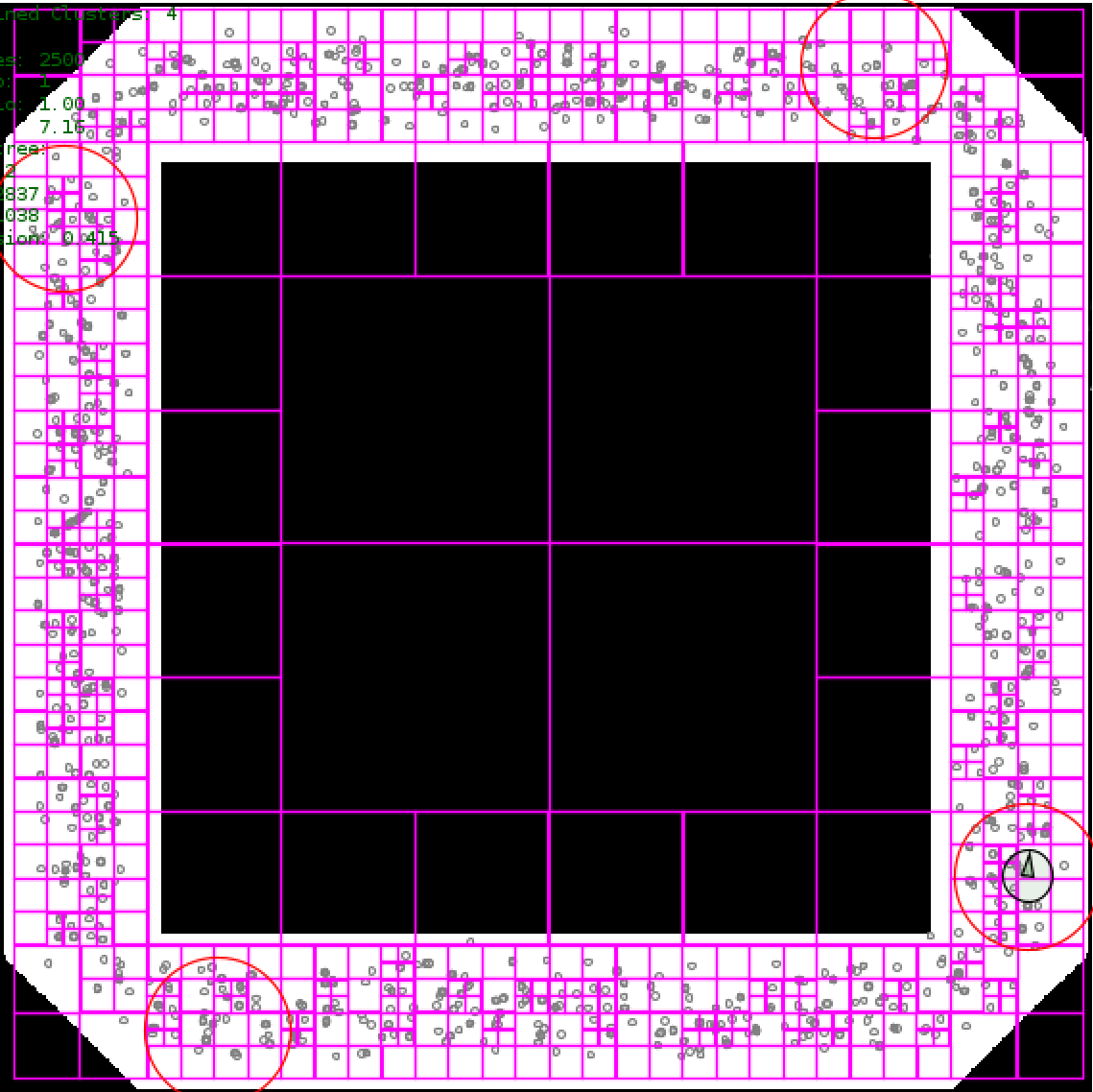}
    }\hspace{0pt}%
    \subfloat[DR-SIR ($t=20$) \label{fig:dr-sir_t20
    }]{
        \includegraphics[width=0.47\columnwidth]{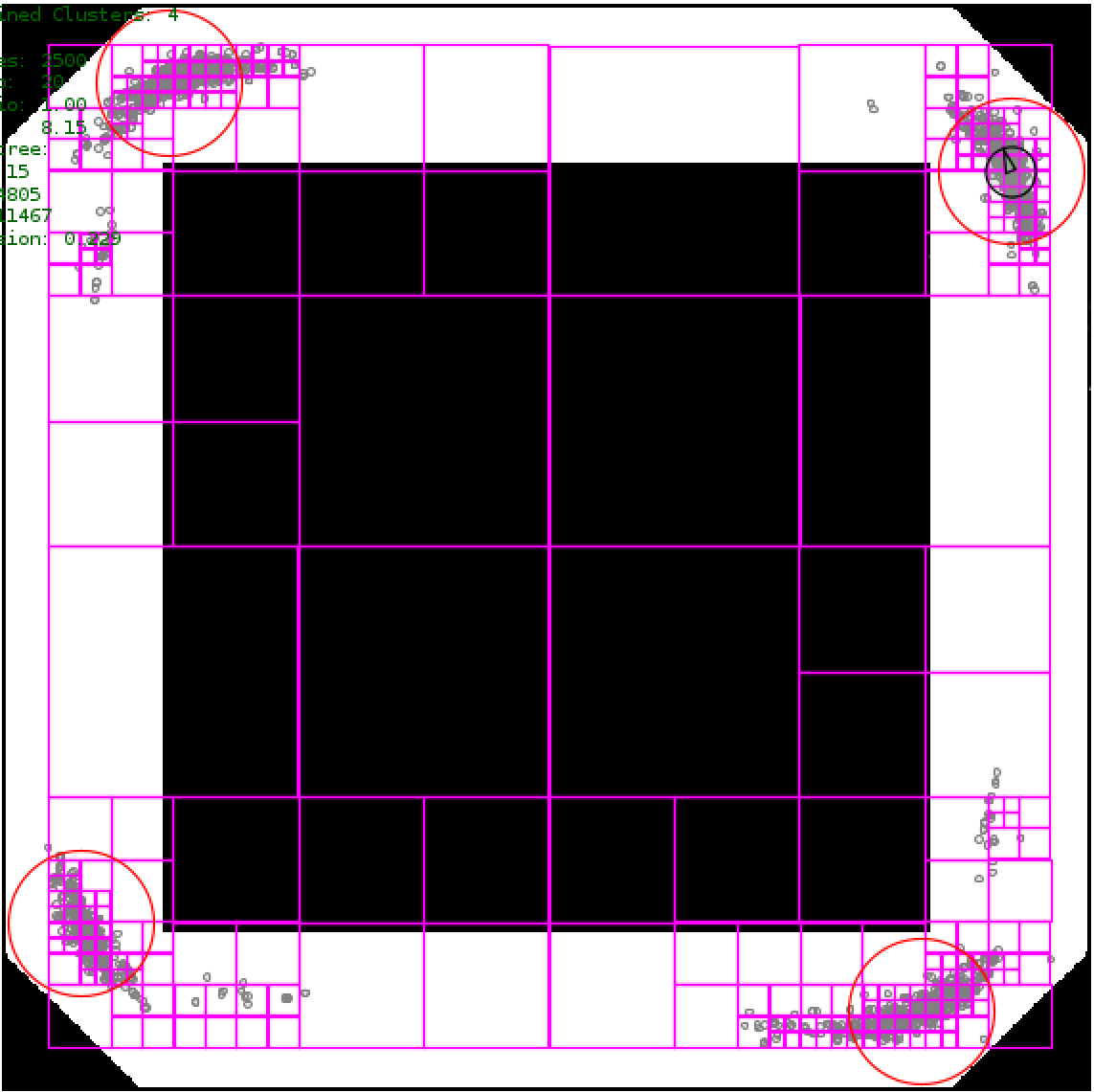}
    }\\[-0.3em]
    % Second row
    \subfloat[PGM-II ($t=0$)\label{fig:pgm_t0}]{
        \includegraphics[width=0.47\columnwidth]{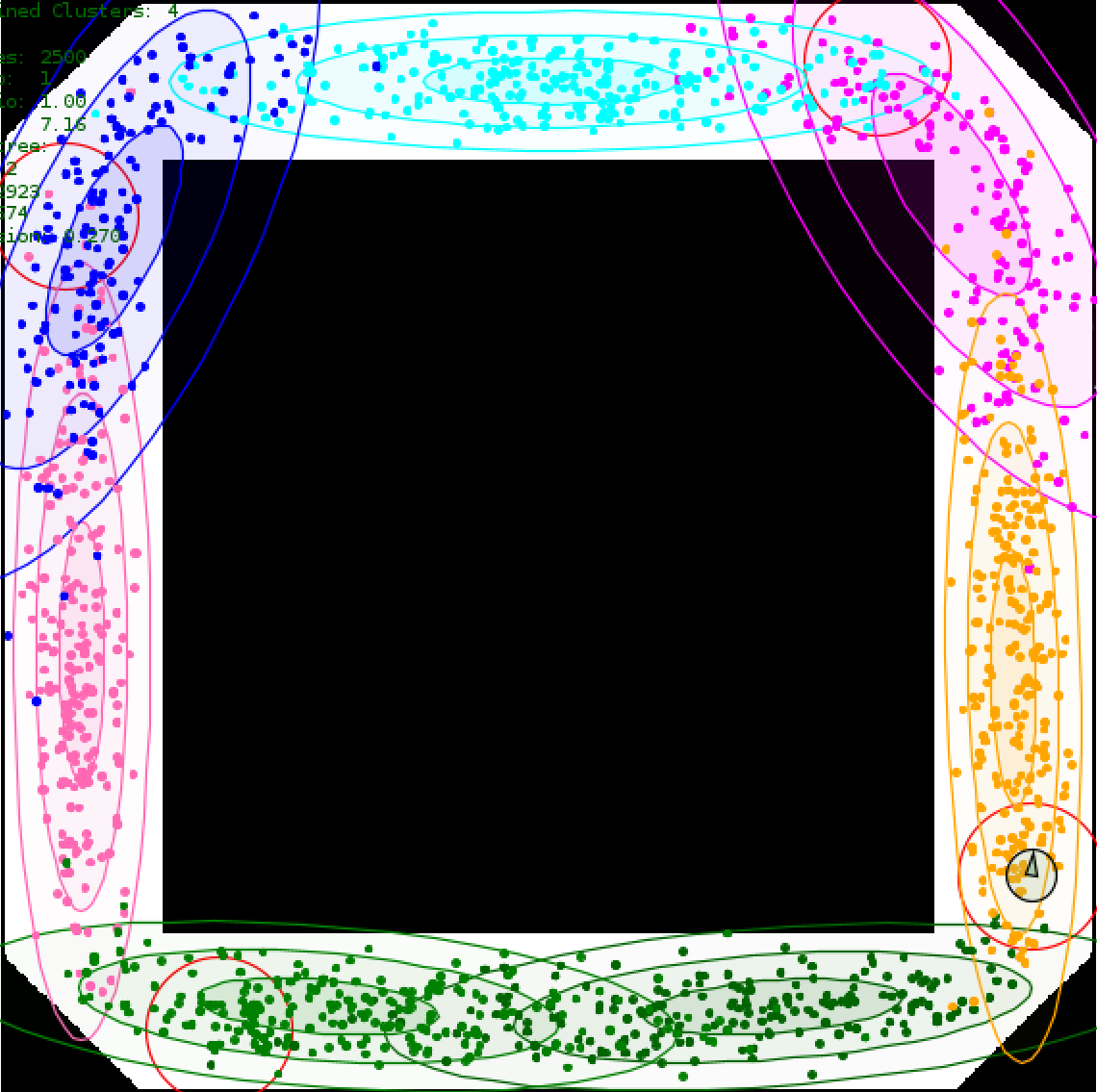}
    }\hspace{0pt}%
    \subfloat[PGM-II ($t=20$)\label{fig:pgm_t20}]{
        \includegraphics[width=0.47\columnwidth]{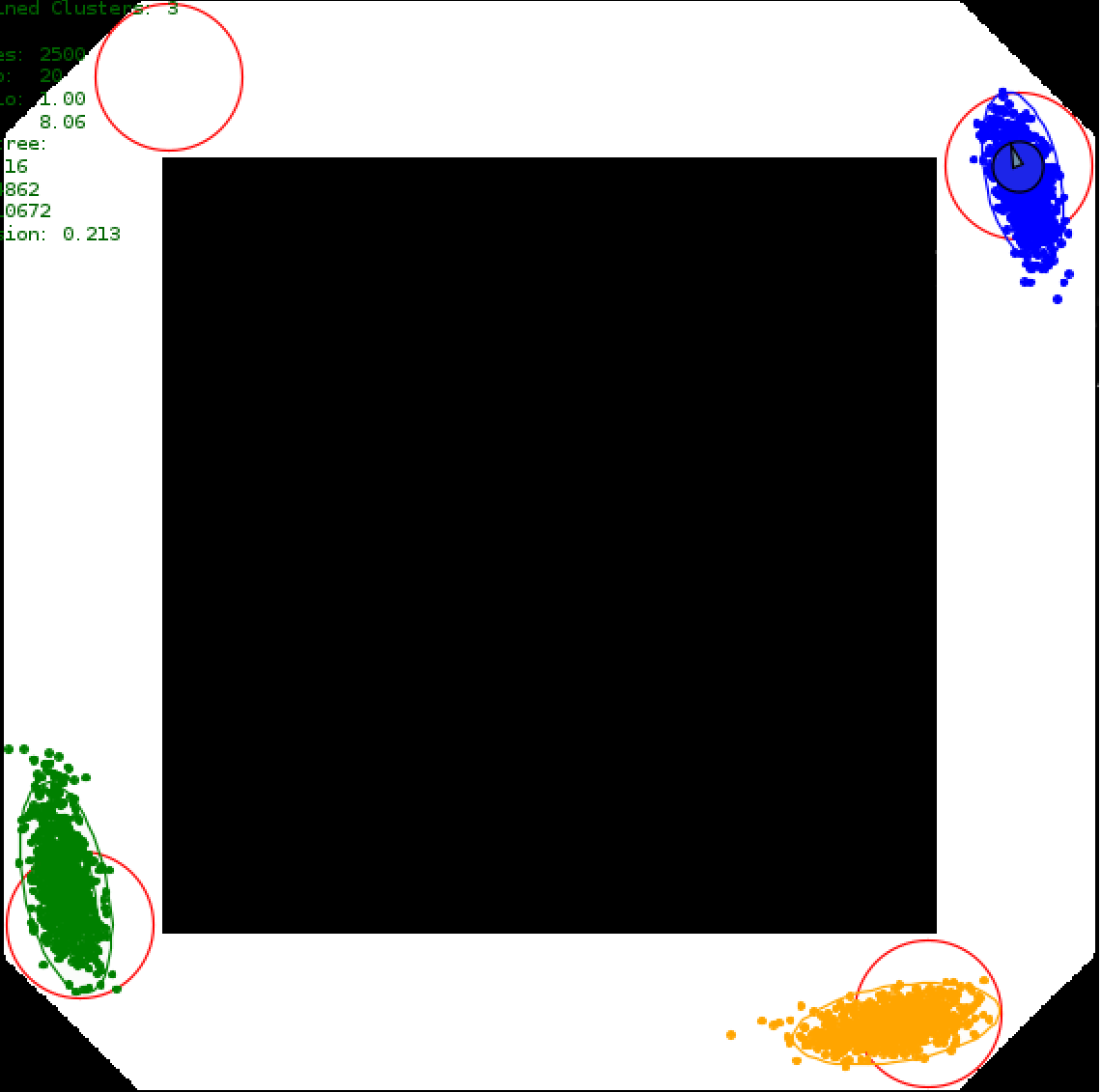}
    }
    \caption{Top row: From start (left), DR-SIR ensures uniform spatial resampling by spatial partitioning (pink cells), converging to the four correct modes (right). Bottom row: PGM-II tries to find optimal clustering from uniform data (left). Often this loses some of the modes very quickly (right).}
    \label{fig:dr-sir_and_pgm_behavior}
\end{figure}

To assess the overall diversity maintenance behavior, we first examine the number of maintained modes over time, for all and for successfully started runs. Figure \ref{fig:maintained_clusters} depicts the average cluster count and its distribution for $P=1000$. The plots are averages over 50 runs with $P=1000$. A run is \textit{started successfully} if it maintains all four modes for the first 50 steps. The difference between all runs (top) and successfully started runs (bottom) highlights sensitivity to initialization. We see that PF consistently fails regardless of initial conditions, and that PGF-II's performance is highly dependent on the early run, where the uniform initialization leads to chaotic Gaussian clustering (see Fig. \ref{fig:pgm_t0}). After surviving the initial stage, all algorithms excluding PF are mostly able to maintain all modes (Fig. \ref{fig:maintained_clusters}, bottom).

As expected, the PF fails to maintain modes, peaking at only 0.30 success rate even at maximum particle count. FDS, DR-SIR, and ATOG variants achieve considerably higher success rates, peaking at 98-100\% at the maximum particle count. PGM-II improves over PF but underperforms. We presume this is due to sensitivity to initialization, failed clustering, and uninformative, non-linear measurements along corridors (Fig. \ref{fig:dr-sir_and_pgm_behavior}). The results are presented in Table \ref{table:kootstra_results} and visualized in Fig. \ref{fig:success_square}.

\subsubsection{Maze environment}
At particle counts up to $P=1000$, FDS, DR-SIR, and ATOG variants similarly outperform PF in success rate. At higher counts, DR-SIR becomes the best performer. Among successful runs, PF achieves the lowest RMSE, outperforming DR-SIR and ATOG variants by approximately 25\%. In contrast, FDS exhibits noticeably higher ($\approx +300\%$) RMSE. See Fig. \ref{fig:success_maze} for the results. PGM-II performs worst. Again, we assume this is due to sensitivity to initialization.

\subsection{Robotics Results Summary}
\begin{figure}[ht]
\centering
\subfloat[][Square (ambiguous)]{
\includegraphics[width=0.98\columnwidth]{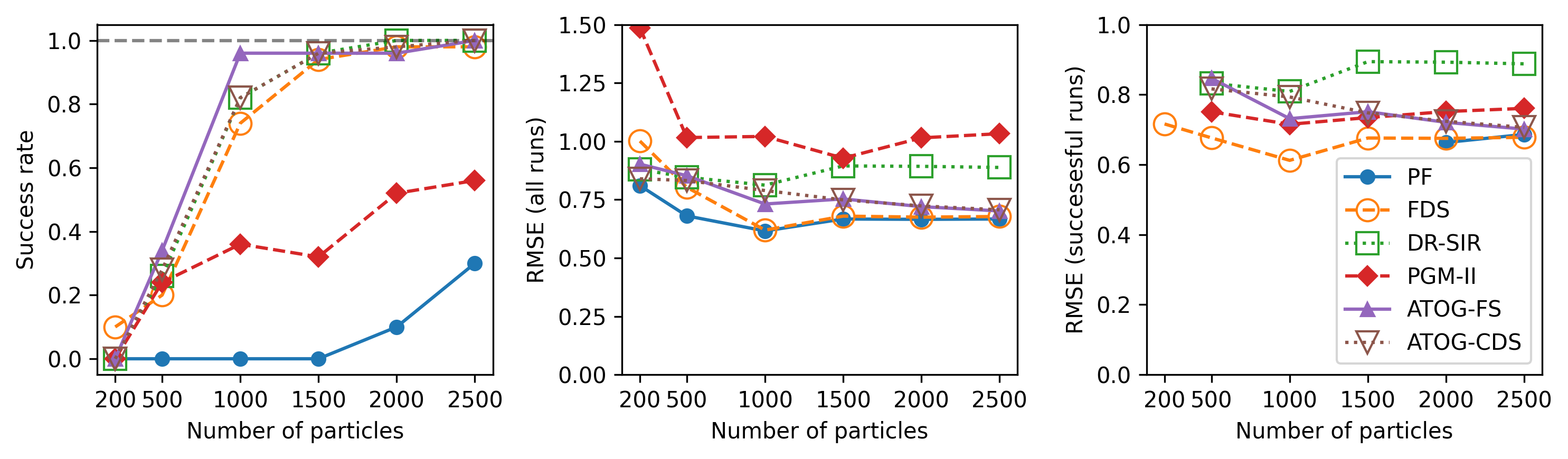}
\label{fig:success_square}}
\qquad
\subfloat[][Maze (non-ambiguous)]{
\includegraphics[width=0.98\columnwidth]{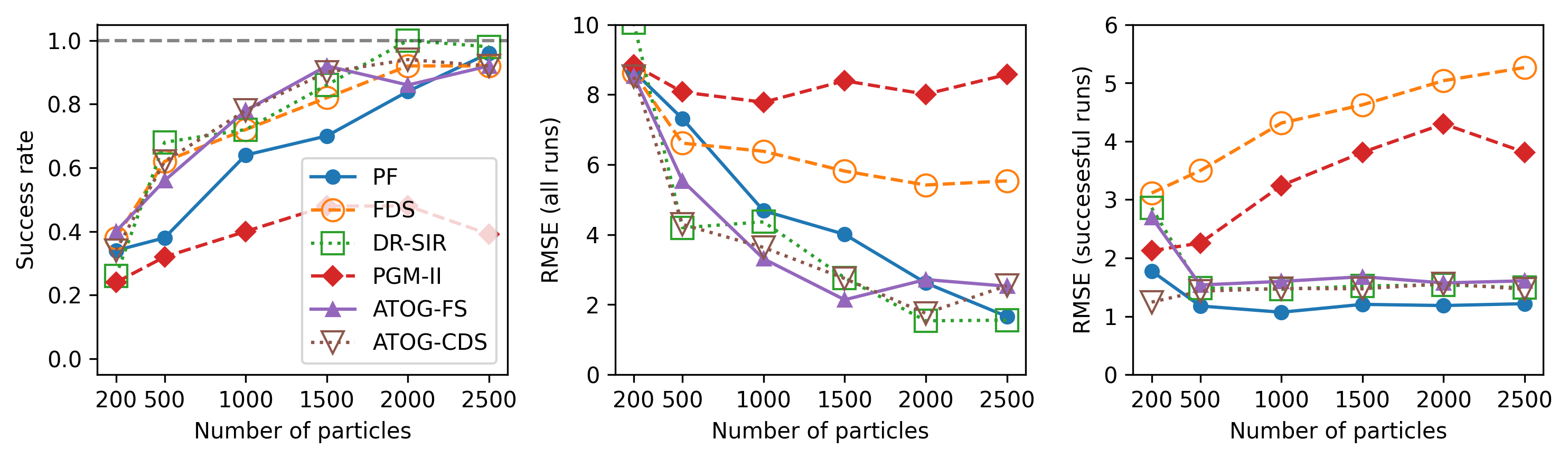}
\label{fig:success_maze}}
\qquad
\caption{Success rates (left), and positioning errors over all (middle) and successful runs (right) over particle counts in both robotics environments (Fig. \ref{fig:environments}). }
\label{fig:success_both}
\end{figure}

Overall, DR-SIR and ATOG variants maintain diversity in the multimodal environment while showing no major weakness in the unimodal case. They trade a small amount of compactness during successful ambiguous runs for high success rates and good overall compactness across both environments (Fig. \ref{fig:success_both}). Based on numerical comparison to implemented algorithms (Table \ref{table:kootstra_results}), ATOG-FS appears at least on par with the best reported diversity maintenance methods without notable drawbacks. 

\begin{table*}[ht]
\scriptsize
% \tiny
\centering
\begin{tabular}{llccccccccc}
\toprule
\multirow{2}{*}{Metric} & \multirow{2}{*}{P} & 
\multicolumn{3}{c}{Kootstra \cite{kootstra2009tackling}} & \multicolumn{6}{c}{Our implementations} \\
\cmidrule(lr){3-5} \cmidrule(lr){6-11}
 & & PF & FDS & Cl. o/t w. &
     PF & FDS & DR-SIR & PGM-II & ATOG-FS & ATOG-CDS \\
\midrule

\multirow{3}{*}{Success rate} 
 & 500  & 0.00*  & 0.15*  & 0.05* & 0.00 & 0.20 & 0.26 & 0.24 & 0.34 & 0.28\\
 & 1000 & 0.00* & 0.45*  & 0.65*  & 0.00 & 0.74 & 0.82 & 0.36 & 0.96 & 0.82 \\
 & 2500 & 0.06 & 0.92 & \bf{1.00} & 0.30 & 0.98 & \bf{1.00} & 0.56 & \bf{1.00} & \bf{1.00} \\
\midrule

\multirow{3}{*}{\makecell{Time to\\premat. conv.}} 
 & 500  & 10* & 80* & 175*          & 42 $\pm$ 35   &   13 $\pm$ 185  & 168 $\pm$ 197      &  185 $\pm$ 198 & 221 $\pm$ 216 & 206 $\pm$ 203 \\
 & 1000 & 25* & 225* & 340*         & 86  $\pm$ 86  &   380 $\pm$ 201 & 419 $\pm$ 171      &  219 $\pm$ 220 & 482 $\pm$ 88 & 417 $\pm$ 176 \\
 & 2500 & 183 & 461 & \bf{500}      & 297 $\pm$ 171 &   490 $\pm$ 63  & \bf{500} $\pm$ 0.0  & 306 $\pm$ 222 & \bf{500} $\pm$ 0.0 & \bf{500} $\pm$ 0.0 \\
\midrule

\multirow{3}{*}{Compactness} 
 & 500  & --     & --    & --       & 0.86 $\pm$ 0.10 & 0.82 $\pm$ 0.10  & 0.78 $\pm$ 0.09 & 0.75 $\pm$ 0.10 & 0.79 $\pm$ 0.11 & 0.80 $\pm$ 0.80 \\
 & 1000 & --     & --    & --       & 0.90 $\pm$ 0.10 & 0.90 $\pm$ 0.10  & 0.81 $\pm$ 0.08 & 0.76 $\pm$ 0.10 & 0.86 $\pm$ 0.12 & 0.82 $\pm$ 0.82 \\
 & 2500 & 0.85*  & 0.8*  & 0.5*     & 0.87 $\pm$ 0.09 & 0.86 $\pm$ 0.08 & 0.89 $\pm$ 0.08  & 0.75 $\pm$ 0.10 & 0.85 $\pm$ 0.10 & 0.85 $\pm$ 0.85 \\
\midrule

\multirow{3}{*}{RMSE} 
 & 500  & -- & --     &  --         & 0.68 $\pm$ 0.37  & 0.80 $\pm$ 0.34  & 0.85 $\pm$ 0.33 & 1.02 $\pm$ 0.30 & 0.85 $\pm$ 0.44 & 0.83 $\pm$ 0.45 \\
 & 1000 & -- & --     & --          & 0.62 $\pm$ 0.36  & 0.62 $\pm$ 0.35  & 0.81 $\pm$ 0.35 & 1.02 $\pm$ 0.36 & 0.73 $\pm$ 0.45 & 0.79 $\pm$ 0.42 \\
 & 2500 & 0.7* & 0.7* & 7.0+*       & 0.67 $\pm$ 0.33  & 0.68 $\pm$ 0.32  & 0.89 $\pm$ 0.32 & 1.03 $\pm$ 0.37 & 0.70 $\pm$ 0.37 & 0.71 $\pm$ 0.46 \\
\bottomrule
\end{tabular}
\caption{This table compares Kootstra's \cite{kootstra2009tackling} and our algorithm implementations across particle counts the Square environment (Fig. \ref{fig:env_square}) Values marked with an asterisk (*) are approximate, visually extracted from graphs in \cite{kootstra2009tackling}. MSSE is converted to RMSE in our units, and a plus sign (+) indicates increasing error over time. Perfect scores are highlighted in bold.}
\label{table:kootstra_results}
\end{table*}

\section{Experiment 2: Application to indoor positioning data}

\begin{figure}[ht]
    \centering
    \subfloat[][PDR data and positioning area]{
        \includegraphics[width=0.68\columnwidth, valign=c]{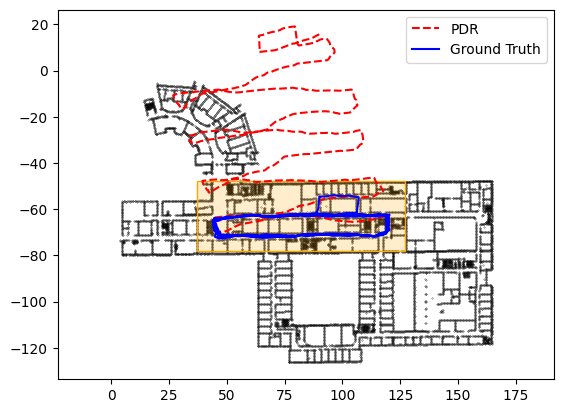}
        \label{fig:maxwell_pdr}
    }
    \subfloat[][Main corridor]{
        \includegraphics[width=0.26\columnwidth, valign=c]{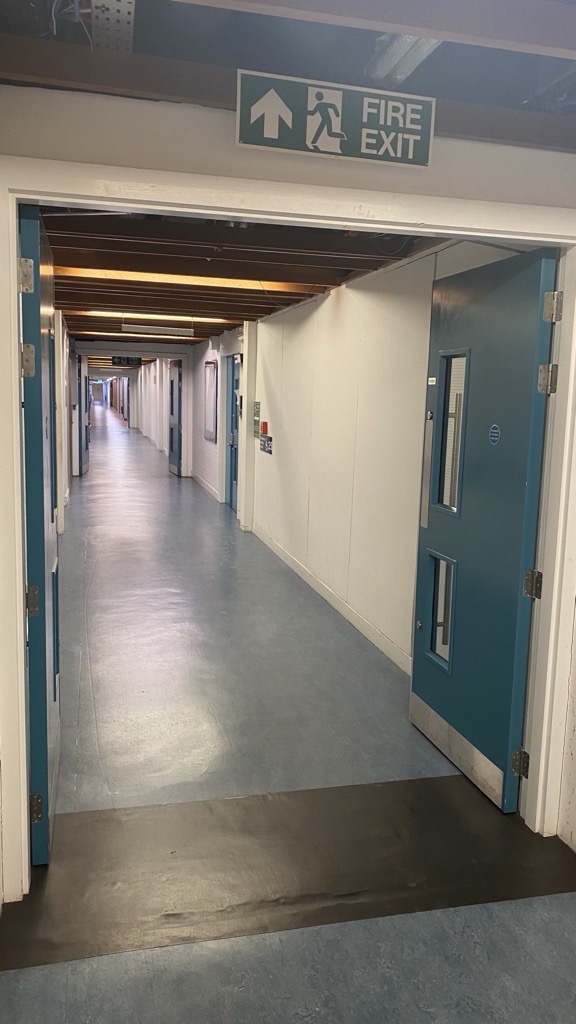}
        \label{fig:maxwell_corridor}
    }
    \caption{James Clerk Maxwell Building}
    \label{fig:maxwell_combined}
\end{figure}

This experiment uses PDR data collected via a Pixel 2 smartphone in James Clerk Maxwell Building, University of Edinburgh, King's Buildings (Fig. \ref{fig:maxwell_pdr}). Data were recorded via the Sensor Logger Android app \cite{sensorlogger_link} and processed into PDR output with a fork of RoNIN ResNet \cite{herath2020ronin} with a pre-trained model (ronin\_resnet.zip). We have made the Sensor Logger-compatible fork publicly available on GitHub \cite{ronin_for_sensorlogger_link}.

The trajectory follows four symmetric loops along the main corridors, finally resolving ambiguity by entering a smaller loop. To highlight differences in diversity maintenance, positioning is restricted to the central floor plan area ($90 \times 30$ \si{\square\metre}). Fig. \ref{fig:maxwell_pdr} visualizes the loops and the restricted area.

We compare five algorithms to evaluate their ability to maintain a bimodal distribution and ultimately converge to the correct path, omitting PGM-II due to limited space, its computational demand, and poor simulation performance.  We use $P = 5000$ particles, and smooth and downsample the trajectory to \SI{0.2}{m} steps. For the motion model we use $\sigma_{tr} = 0.2$, a cross-term $\sigma_{rot\_tr} = 0.06$ (per \SI{1}{m} traveled), and $\sigma_{rot} = 0.3$ (per $2\pi$ rotation). Particle weights are computed via simple floor plan matching \cite{davidson2010application, zampella2015indoor, kaiser2011human}, where crossing a wall is heavily penalized and staying in the middle of the corridor is slightly rewarded.

We compute the final trajectory as the average over all particle histories. Performance is measured by the RMSE of the final trajectory and a success rate, where a run is considered \textit{successful} if  the RMSE is below \SI{15}{m}, i.e., if the filter avoids premature convergence to an incorrect mode. Somewhat random minor deviations along sub-paths, caused by environmental multimodality beyond the main symmetric modes, slightly influence RMSE in successful runs (Fig. \ref{fig:maxwell_runs}, third from top).

\subsection{Results: Indoor Positioning Experiment}
\begin{figure}[ht]
    \centering
    \subfloat[][CDF for RMSE (m) over all and successful runs. The x-axis is in log scale.]{
        \includegraphics[width=0.6\columnwidth, valign=c]{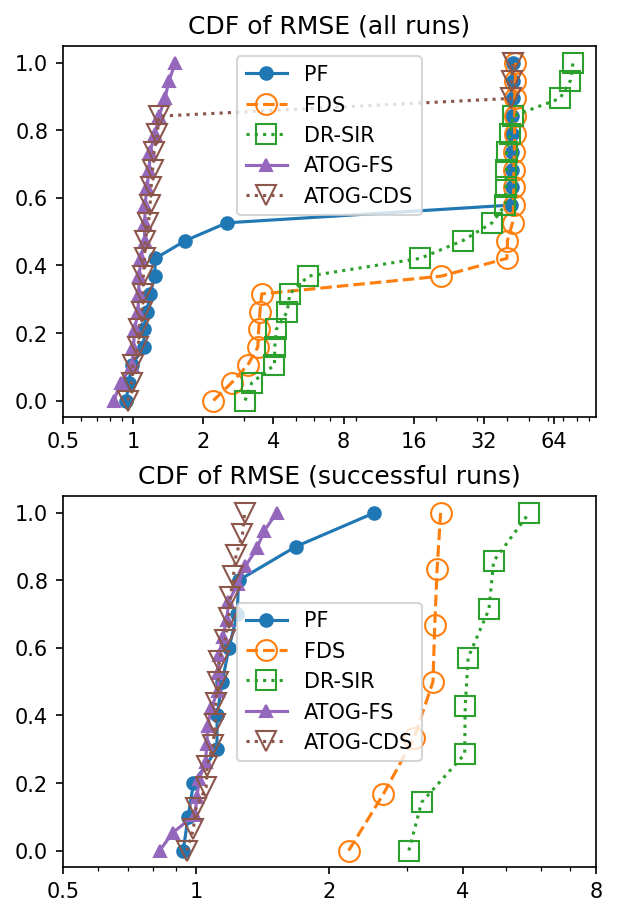}
        \label{fig:maxwell_cdf}
    }
    \subfloat[][Example runs]{
        \includegraphics[width=0.32\columnwidth, valign=c]{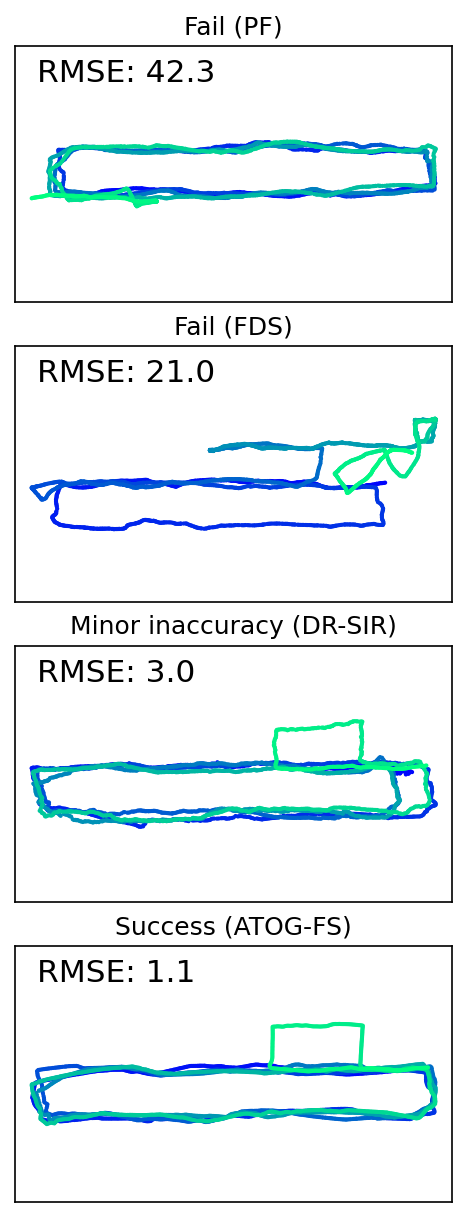}
        \label{fig:maxwell_runs}
    }
    \caption{Indoor Positioning Errors}
    \label{fig:maxwell_errors}
\end{figure}

PF is expected to fail in 50\% of the runs, reflected in its 11/20 success rate. Surprisingly, both FDS and DR-SIR completely collapse in this environment, succeeding only 7/20 and 8/20 times, respectively, and having the highest RMSE even in successful runs. We suspect that their artificial exploratory nature sacrifices too many resources in the small doorways and corridors and/or that their 2D clustering/density does not capture the true bidirectional modes. (We explore the reasons briefly in Appendix \ref{appendix:bimodal_tracking}.) ATOG-FS achieves a perfect 20/20 success rate, while the simpler variant ATOG-CDS succeeds in 17/20 runs. Among successful runs, ATOG-FS and ATOG-CDS achieve the lowest RMSE (\SI{1.1}{m}), closely followed by PF (\SI{1.3}{m}). Overall, the ATOG variants dominate in this environment. Figure \ref{fig:maxwell_errors} shows the cumulative error distributions and example trajectories. Table \ref{table:indoor_positioning_results} summarizes the results.

\begin{table}[ht]
\tiny
\setlength{\tabcolsep}{4pt}
\centering
\begin{tabular}{lccccc}
\hline
 & PF & FDS & DR-SIR & ATOG-FS & ATOG-CDS (\ref{eq:cluster_dependent_selection}) \\
\hline
Success rate       &   0.55 (11/20)         & 0.35 (7/20)       & 0.40 (8/20) & $\mathbf{1.00}$ (20/20)  & 0.85 (17/20)  \\
Minor inaccuracy   &   $0.27$ (3/11)        & 0.71 (5/7)        & 0.63 (5/8) & $\mathbf{0.0}$ (0/20)    & 0.0 (0/17)  \\
RMSE (all)         &   $19.7 \pm 9.5$       & $27.8 \pm 12.4$   & $28.7 \pm 8.7$ & $\mathbf{1.1} \pm 0.46$ & $7.3 \pm 3.1$      \\
RMSE (success)     &$  1.3 \pm 0.51 $       & $3.1 \pm 1.0$     & $4.2 \pm 5.2$ & $\mathbf{1.1} \pm 0.46$ & $\mathbf{1.1} \pm 0.40$   \\
\hline
\end{tabular}
\caption{Indoor Positioning Experiment results: RMSE with standard deviations.}
\label{table:indoor_positioning_results}
\end{table}

\section{Conclusions, Limitations, and Future Work}
We have presented an alternative to distance-based clustering for PF diversity maintenance. Its novelty lies in using ancestry tree topology to form clusters that reflect similarity in a way that adapts to the modal nature of the environment, without major effect on compactness. Its simplicity is a key strength: while maintaining the ancestry tree requires some work, ATOG then offers an easy way to control particle diversity. Even the simplest modification to the standard PF (ATOG-CDS) is notably more resilient to multimodality. 

Unlike many domain-tuned algorithms, ATOG demonstrates robustness across diverse environments. When FDS loses compactness with unimodality, and DR-SIR shines in simple environments but completely collapses in a real-world scenario, ATOG consistently performs well without domain-specific tuning. Although the environments and trajectories are intentionally toy-like for clarity, the results are encouraging. Most importantly, we provide a conceptual starting point for future refinements, such as: 
\begin{enumerate}[label=(\alph*)]
    \item Adaptive cluster size and definition (\ref{eq:tree_weighs}--\ref{eq:tree_clusters}) and weighing strategies beyond a constant factor (e.g., by cluster size) (\ref{eq:cluster_dependent_selection}) to enable different behaviors
    \item Multi-level clusters (major and minor) for more granular diversity control and convergence analysis
    \item Combining with other approaches
\end{enumerate}

Potential future work includes:
\subsubsection{\textbf{High-dimensional problems}} Unlike many other clustering algorithms, ATOG is not dependent on problem dimension, making evaluation on high-dimensional problems an interesting direction.

\subsubsection{\textbf{Application to unbalanced environments}}
This work was partly motivated by geometrically unbalanced indoor environments, where diversity is lost due to differences in corridor width. Such issues have been addressed with geometry-based mobility models \cite{kaiser2011human}. We expect ATOG to be applicable here as well, potentially reducing or removing the need for dedicated mobility models or geometric heuristics.

\subsubsection{\textbf{Cluster uses beyond niching}}
One immediate use for the clusters is a more efficient variant of \textit{Kernel Density Estimation}, where summation could be performed over cluster leaves rather than all particles. This may also be more appropriate in situations where (overlapping) spatial closeness does not fully capture similarity, e.g., bidirectional flows (Fig.~\ref{fig:ancestry_tree_pos} and \ref{fig:pgm_t0}).

\bibliographystyle{elsarticle-num} 
\bibliography{bibliography}

\appendices

\section{Bimodal tracking in the Campus Building}
\label{appendix:bimodal_tracking}
We were genuinely surprised by the performance collapse of FDS and DR-SIR in the indoor positioning experiment, so we decided to investigate further to find possible reasons for the failure. To make the behavior clearer, we focused on tracking instead of global positioning (followed by tracking). For this, we initialized the particles around the two symmetric modes and reduced the particle count from 5000 to 1000 to make failure more likely. We found some likely explanations for the poor performance. A video showcasing this experiment is available at \url{https://youtu.be/VkeF1HAEneU}.

\subsection{FDS - why it fails in the indoor environment?}
\begin{figure}[h!]
\centering
\fbox{\includegraphics[width=0.95\columnwidth]{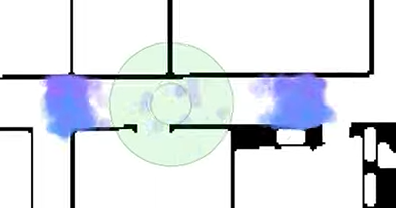}}
\caption{As FDS penalizes particles in dense areas, it often makes the cloud split into two. This possibly loses track of the correct estimate (green). This is highlighted in long corridors without geometric constrains in the direction of travel. Here, both subclouds are traveling from left to right (color depicts the direction).}
\label{fig:fds_fail_split}
\end{figure}

In contrast to the robotics environment, the campus building has long segments without discriminating geometric features in the direction of travel. The filter must maintain a relatively compact estimate based solely on PDR. This is where FDS struggles: it assumes the cloud is too dense around the true position and artificially splits it by favoring particles without spatially close neighbors. In some cases, this completely loses the estimate (or mode). We suspect similar reasons for the failure in the positioning experiment. Figure \ref{fig:fds_fail_split} illustrates this behavior.

\subsection{DR-SIR - why it fails in the indoor environment?}
\begin{figure}[htbp]
  \centering
  \subfloat {
    \centering
    \fbox{\includegraphics[width=\linewidth]{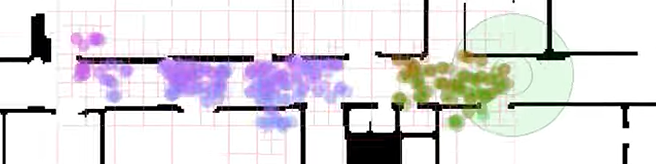}}
  } \\
  \subfloat{
    \centering
    \fbox{\includegraphics[width=\linewidth]{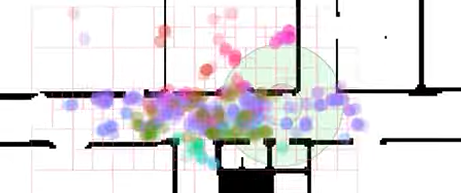}}
  } \\
  \subfloat {
    \centering
    \fbox{\includegraphics[width=\linewidth]{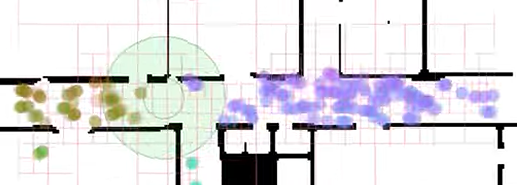}}
  }
  \caption{Bidirectional flows break DR-SIR: As we use only 2D coordinates for spatial binning, colliding subclouds (top) create badly-behaving support particles that explode into random directions (middle). This leaves the correct subcloud badly damaged (bottom).}
  \label{fig:dr-sir_particle_collision}
\end{figure}

DR-SIR aims to maintain good spatial coverage and tends to stretch the particle cloud along long corridors. This alone does not cause DR-SIR to fail in this experiment. However, because we bin the particles using only their 2D coordinates, we observe an issue with bidirectional flows: colliding subclouds create poorly behaving support particles \cite{li2012deterministic}. Although this is rare in this tracking scenario, we suspect that in the initial stages of positioning it is the main reason for the poor performance. This effect is visualized in Figure \ref{fig:dr-sir_particle_collision}.

Naturally, this issue might be fixed with different similarity measures and binning strategies, such as a hand-tailored spatial search tree on \(SE(2)\). However, this further underscores the difficulty of defining domain-specific similarities and the potential computational costs associated with them.

\section{Robotics simulation visualized}
To help the reader understand the characteristics of the different algorithms, we visualize selected snapshots from the robotics experiments in Figures \ref{fig:app_pf}–\ref{fig:app_atog}. The timestamps are approximately $t=\{0,20,50,400\}$, shown from left to right. A video version of this comparison is available at \url{https://youtu.be/8c55IgosWCs}

\begin{figure*}[t]
  \centering

  \newcommand{\SubFigImg}[1]{%
      \includegraphics[trim=10 10 10 10,clip, width=0.23\textwidth]{figures/appendix_robot_square/#1}%
  }

  \newcommand{\FourImageRow}[6]{%
    \begin{minipage}{\textwidth}
        \centering
        \SubFigImg{#1}\hfill
        \SubFigImg{#2}\hfill
        \SubFigImg{#3}\hfill
        \SubFigImg{#4}%
        \captionsetup{skip=1pt}
        \caption{#5}
        \label{#6}
    \end{minipage}\par\vspace{6pt}
  }

  \FourImageRow{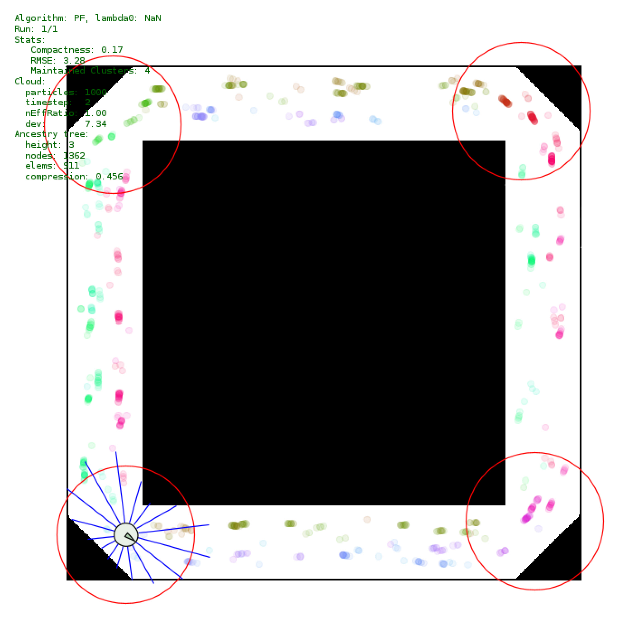}{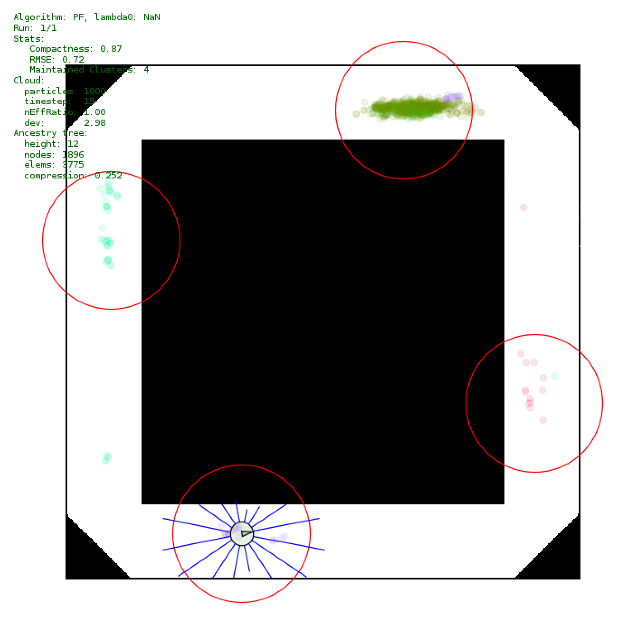}{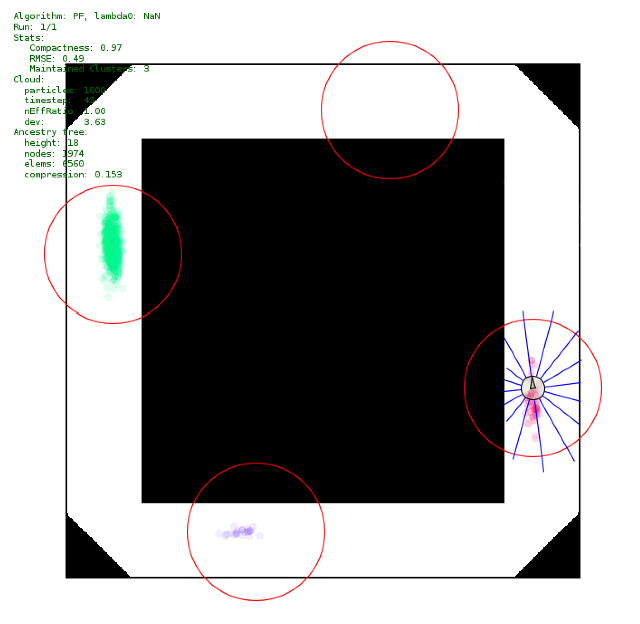}{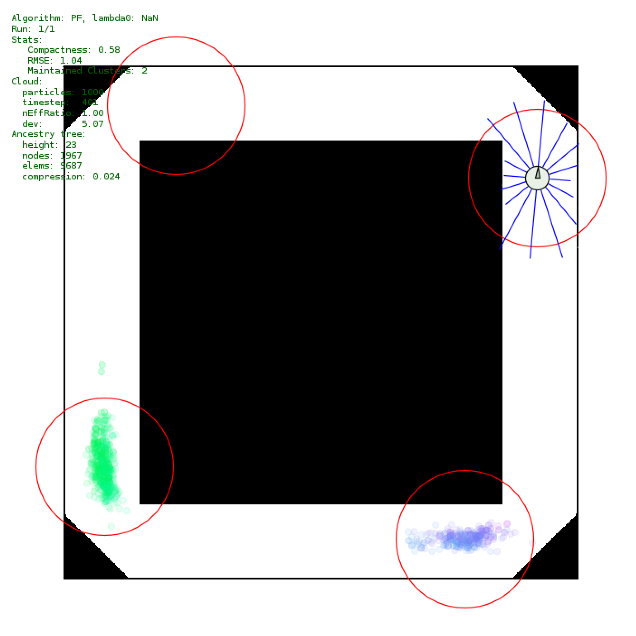}{Standard PF converges prematurely and loses some of the modes very quickly. In this run, it retains only two of the modes.}{fig:app_pf}
  \FourImageRow{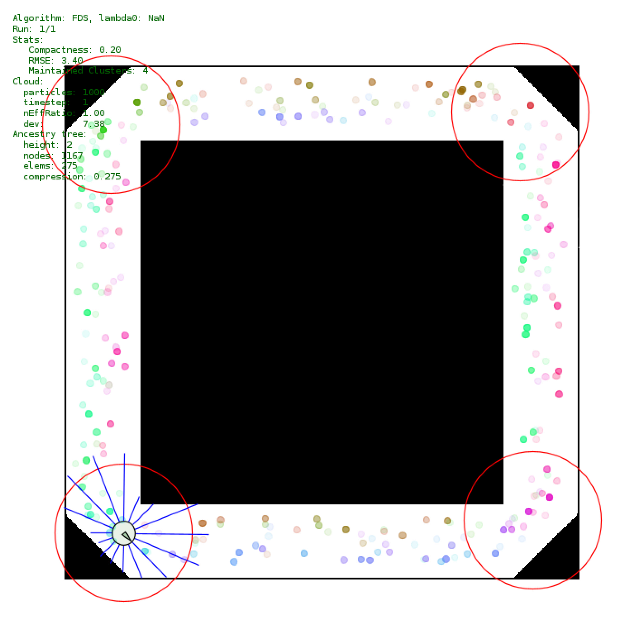}{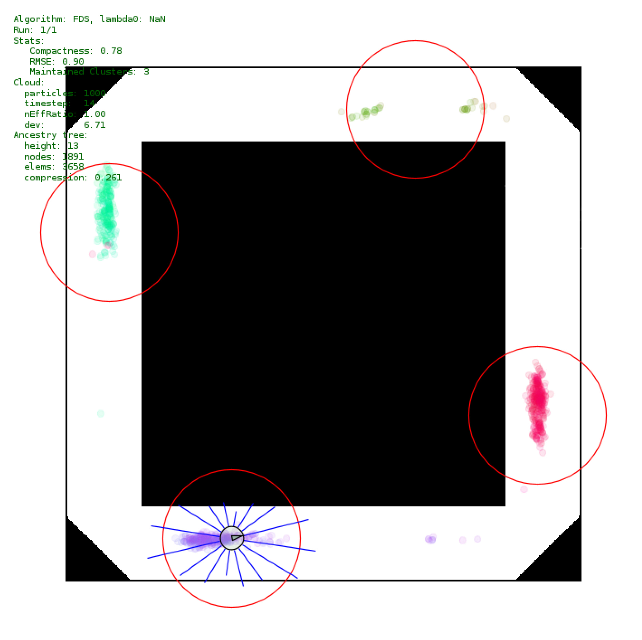}{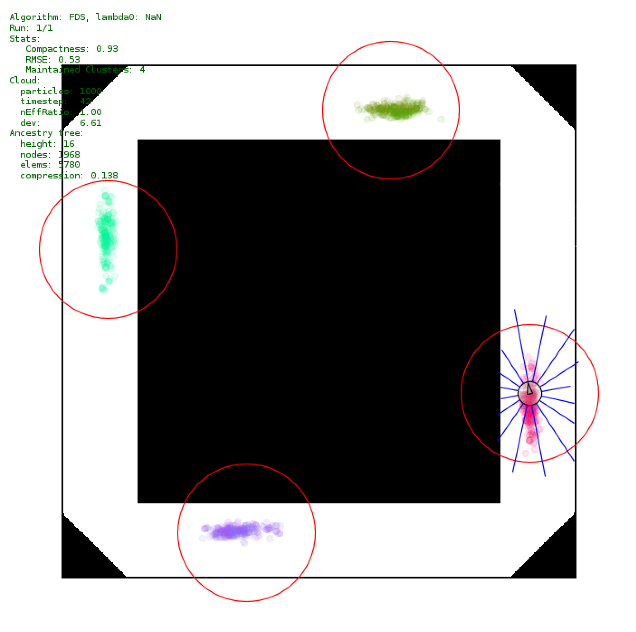}{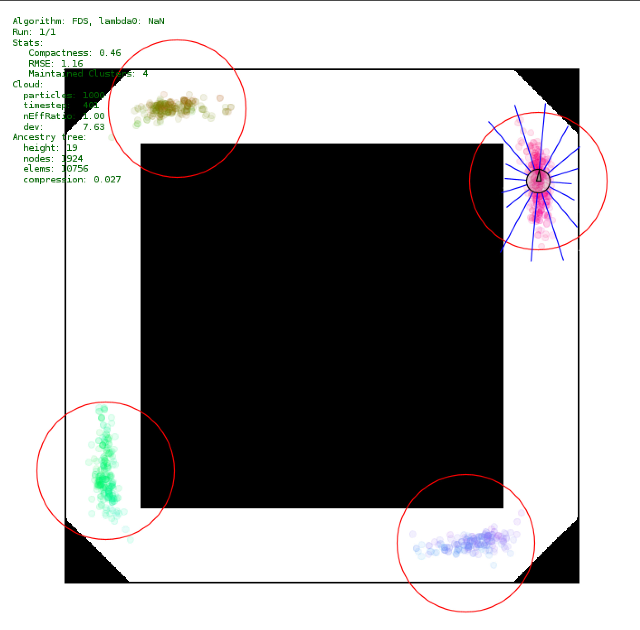}{Frequency-dependent selection (FDS) is able to maintain all four modes, despite temporally splitting one of them (second from left).}{fig:app_fds}
  \FourImageRow{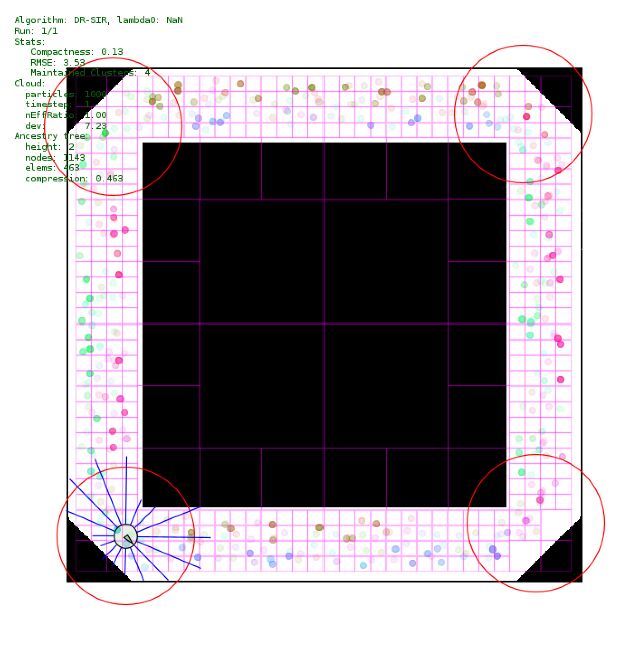}{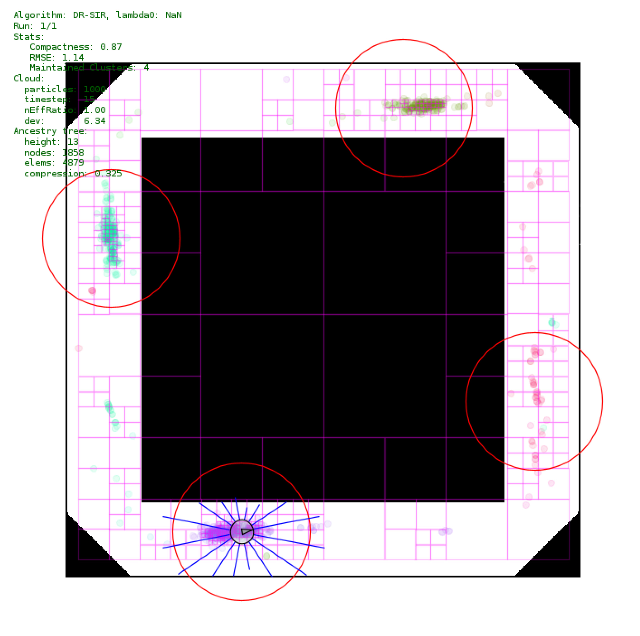}{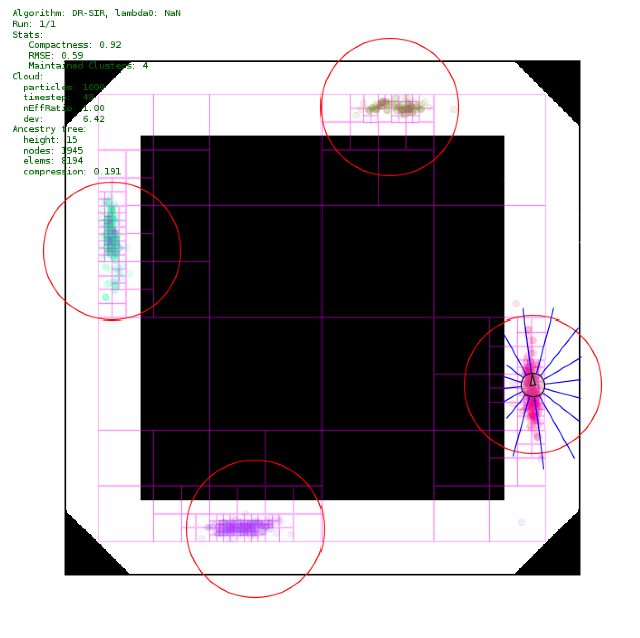}{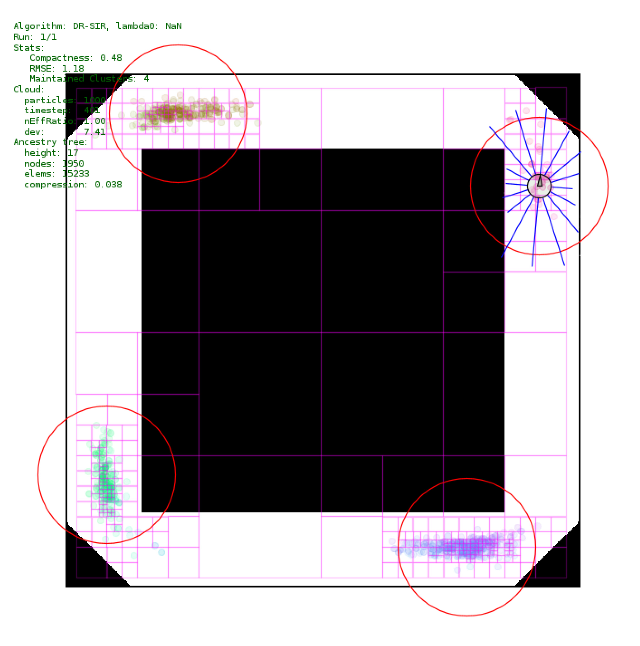}{Deterministic Resampling (DR-SIR) is able to maintain all four modes without problems.}{fig:app_dr-sir}

  \label{fig:robotics_visuals_square_1}
\end{figure*}

\begin{figure*}[t]
  \centering

  \newcommand{\SubFigImg}[1]{%
      \includegraphics[trim=10 10 10 10, clip, width=0.23\textwidth]{figures/appendix_robot_square/#1}%
  }

  \newcommand{\FourImageRow}[6]{%
    \begin{minipage}{\textwidth}
        \centering
        \SubFigImg{#1}\hfill
        \SubFigImg{#2}\hfill
        \SubFigImg{#3}\hfill
        \SubFigImg{#4}%
        \captionsetup{skip=1pt}
        \caption{#5}
        \label{#6}
    \end{minipage}\par\vspace{6pt}
  }

  \FourImageRow{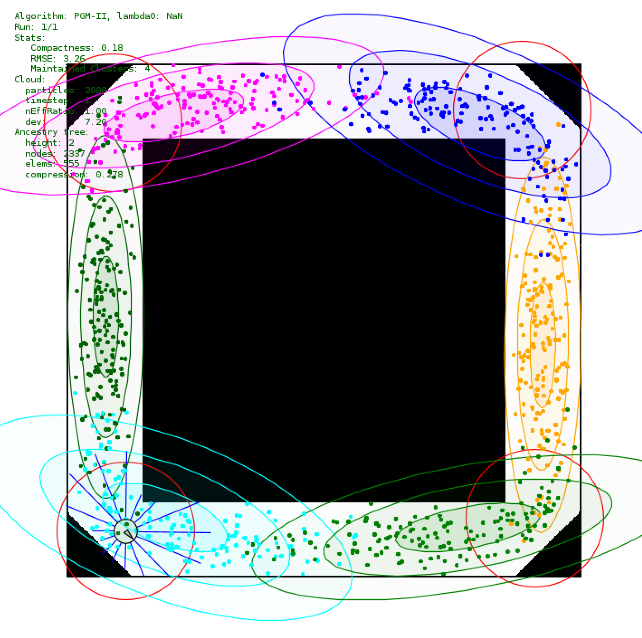}{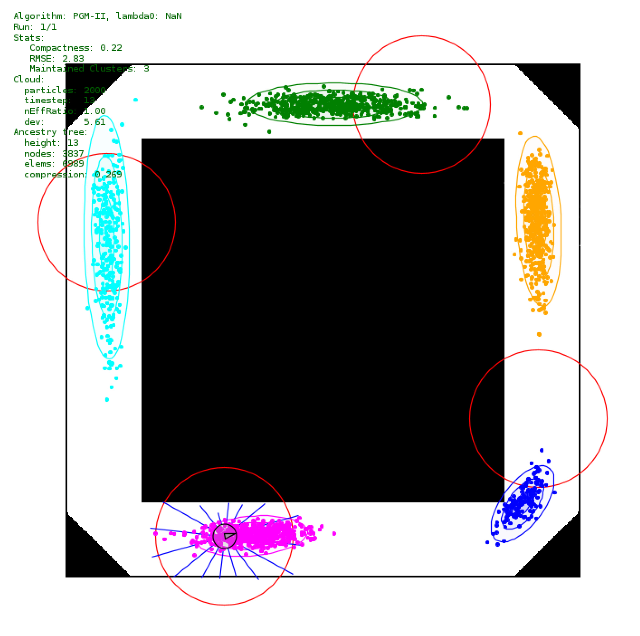}{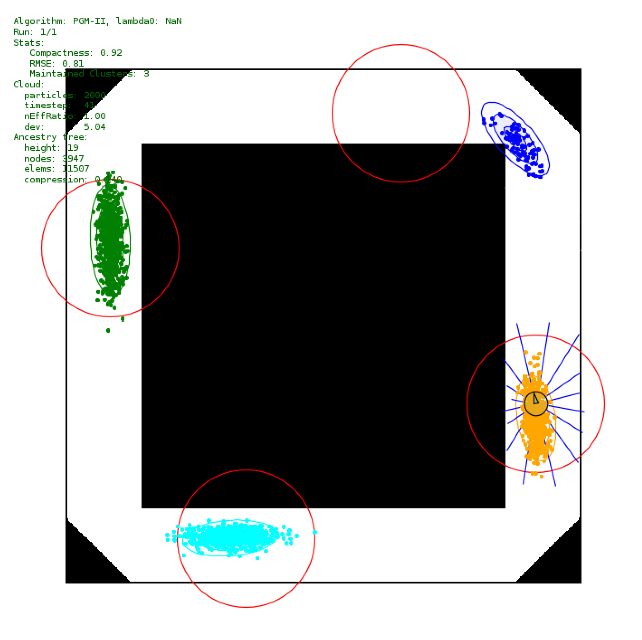}{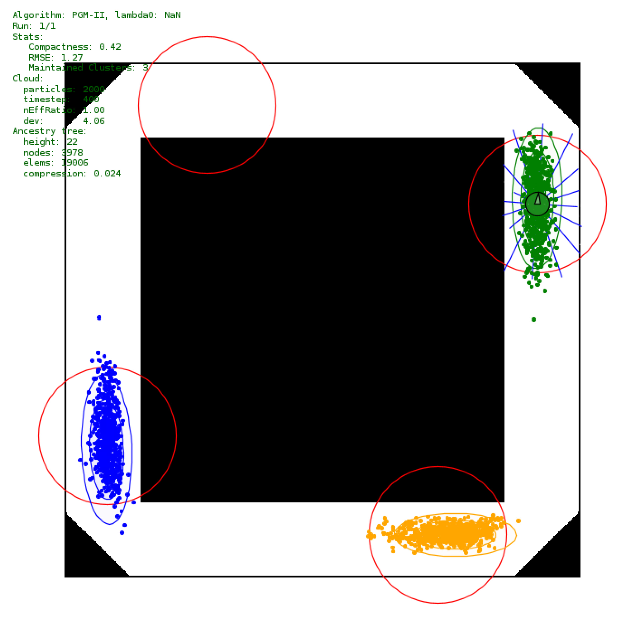}{PGM-II suffers from chaotic initial clustering. Different modes are sometimes wrongly clustered together, and this leads to incorrect Gaussians to sample from. In this run, it is able to maintain three our of four modes. The color here depicts Gaussian cluster index.}{fig:app_pgm}
  \FourImageRow{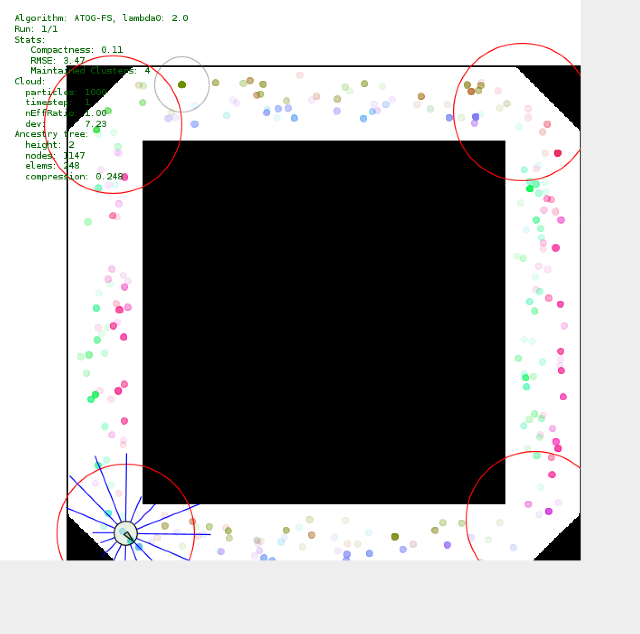}{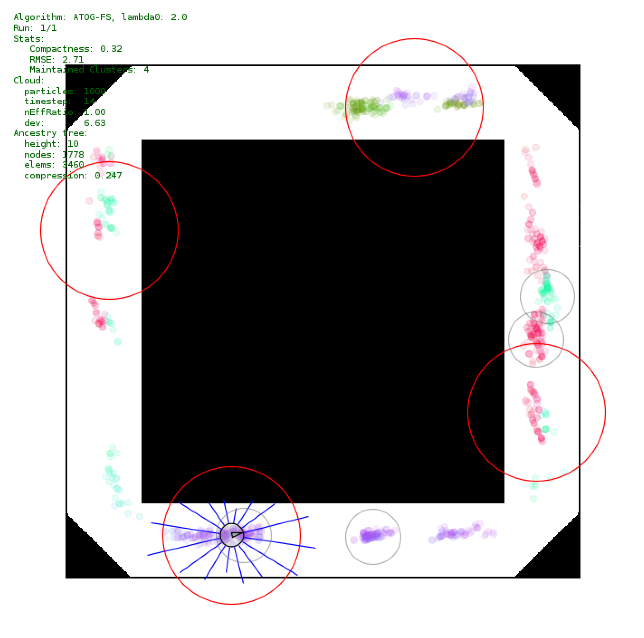}{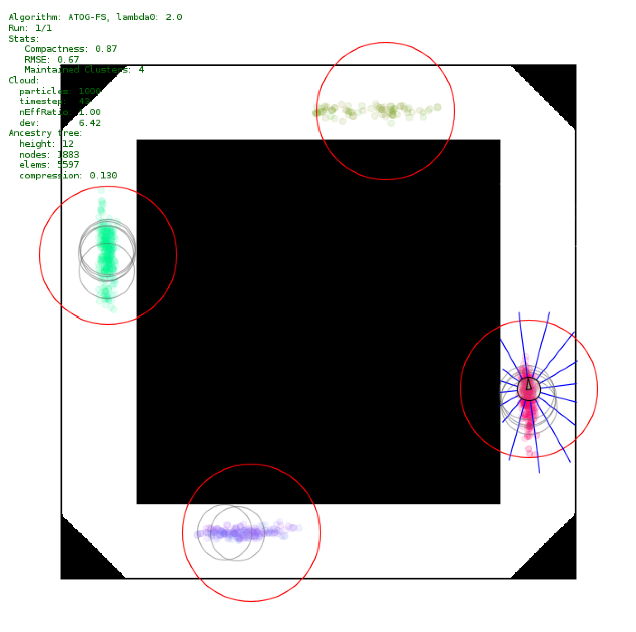}{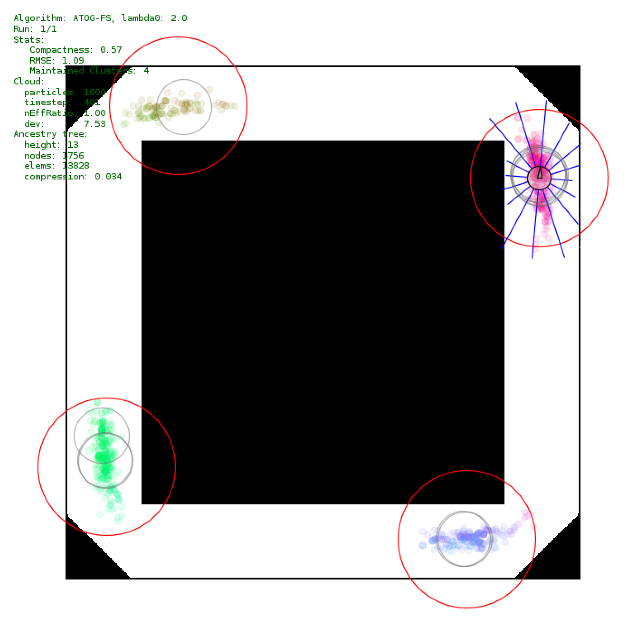}{ATOG-FS is able to maintain all four modes without problems.}{fig:app_atog}

  \label{fig:robotics_visuals_square_2}
\end{figure*}

\begin{figure*}[t]
  \centering

  \newcommand{\SubFigImg}[1]{%
      \includegraphics[trim=10 10 10 10,clip, width=0.23\textwidth]{figures/appendix_robot_maze/#1}%
  }

  \newcommand{\FourImageRow}[5]{%
    \begin{minipage}{\textwidth}
        \centering
        \SubFigImg{#1}\hfill
        \SubFigImg{#2}\hfill
        \SubFigImg{#3}\hfill
        \SubFigImg{#4}%
        \captionsetup{skip=1pt}
        \caption{#5} % row-level caption (unnumbered)
    \end{minipage}\par\vspace{6pt}
  }

  \FourImageRow{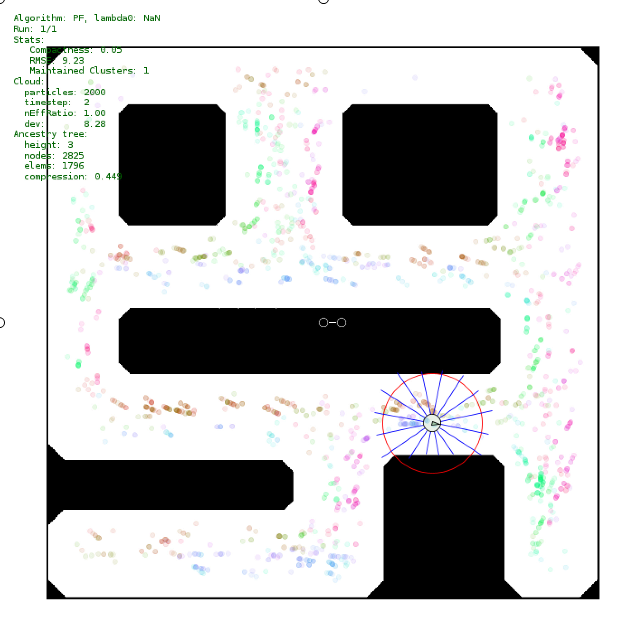}{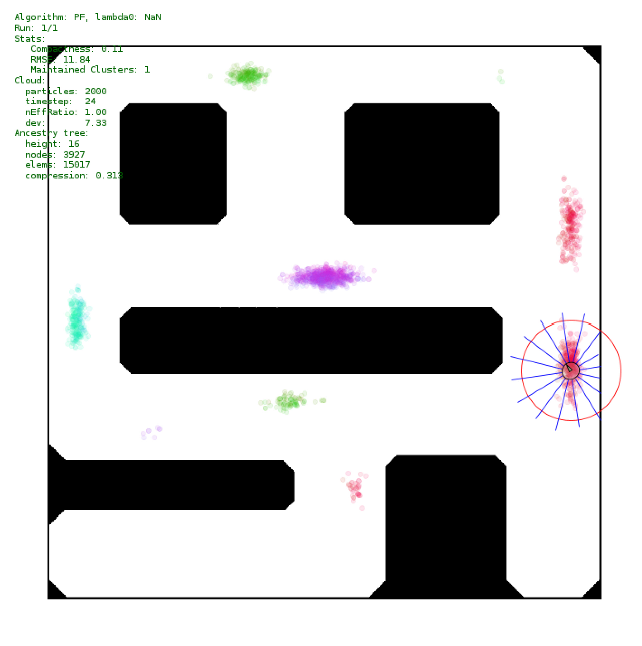}{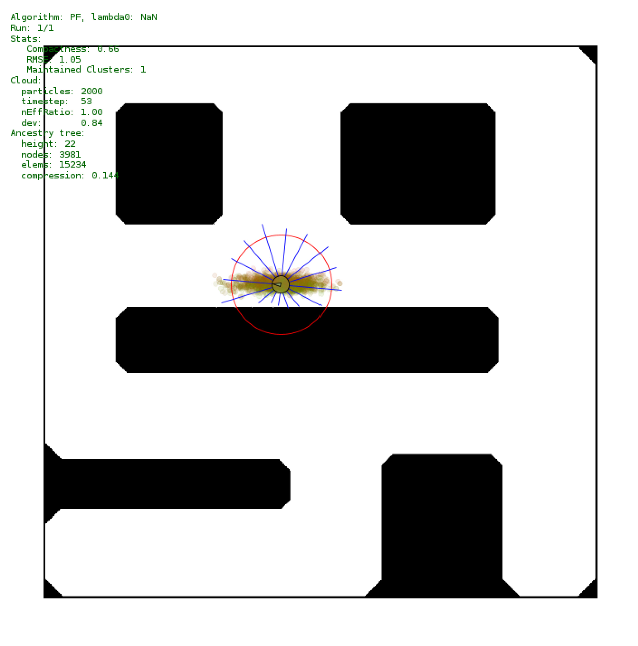}{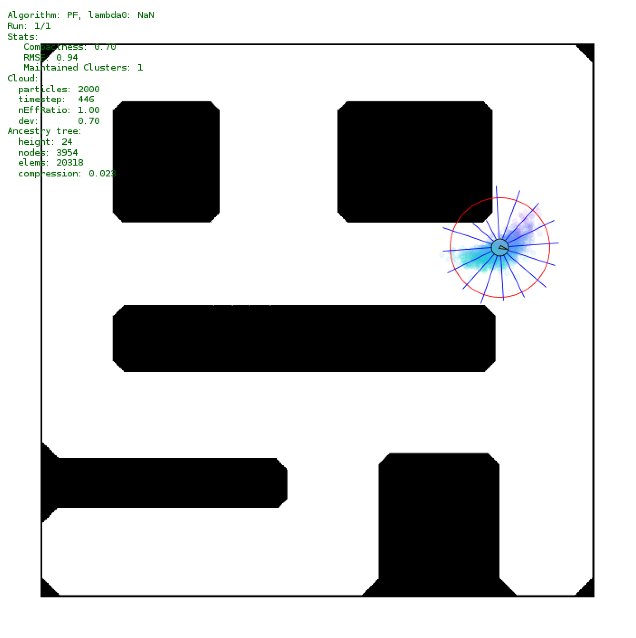}{Standard PF performs well in the unimodal environment. It converges fast and keeps the cloud compact around the correct position.}
  \FourImageRow{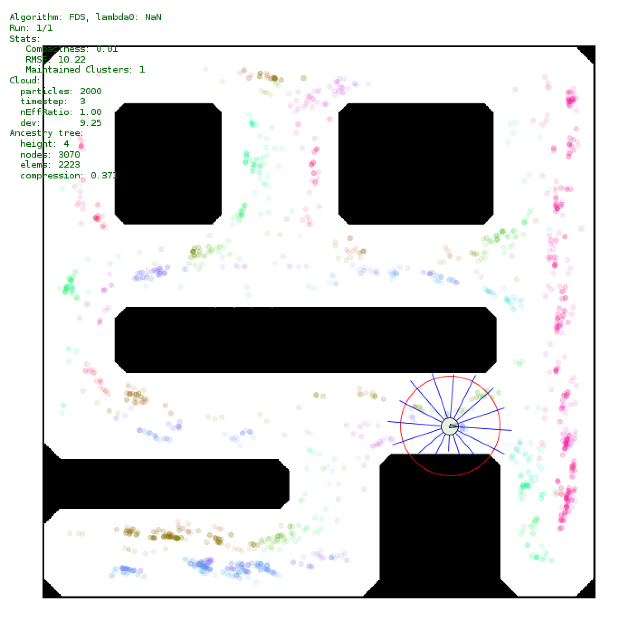}{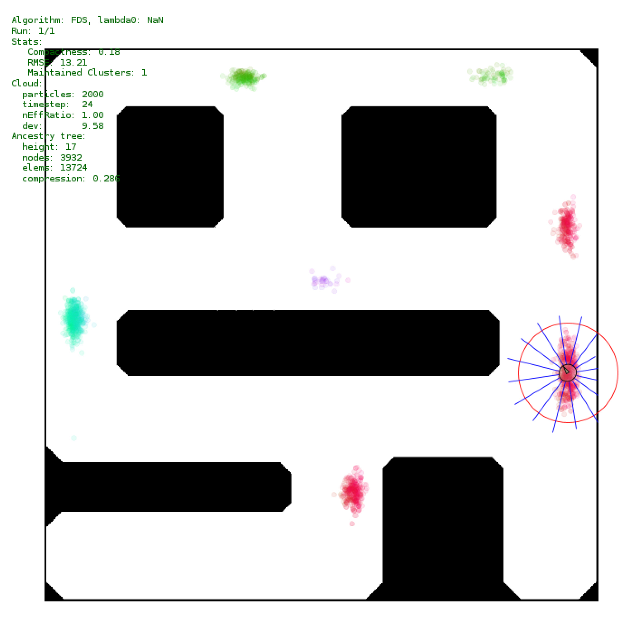}{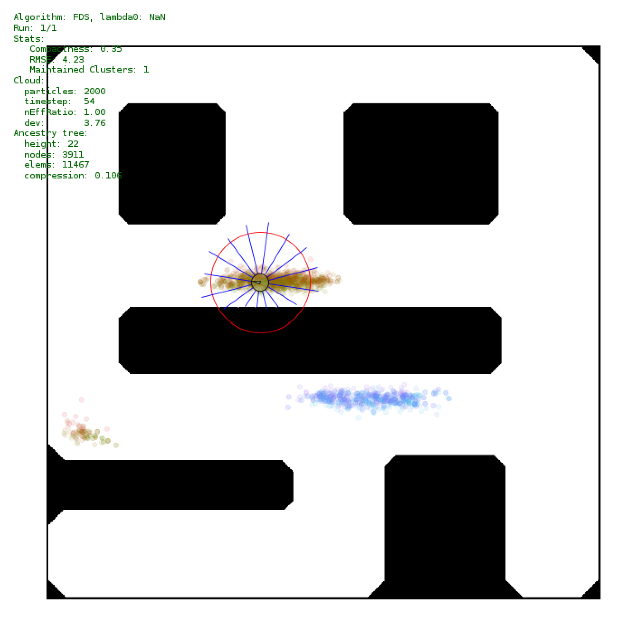}{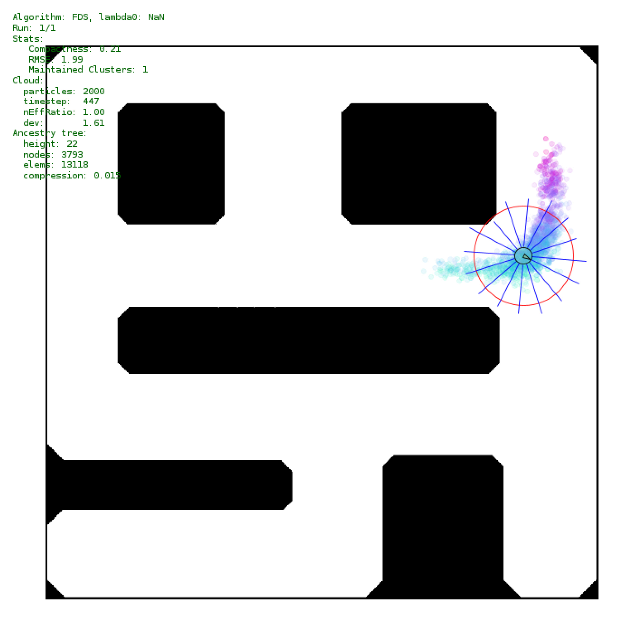}{Frequency-dependent selection (FDS) retains multiple modes for a while. After convergence it struggles to keep the estimate compact (right). }
  \FourImageRow{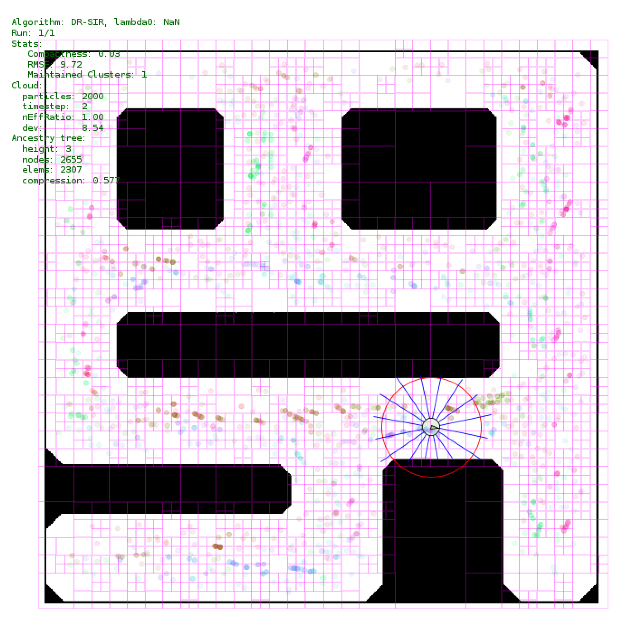}{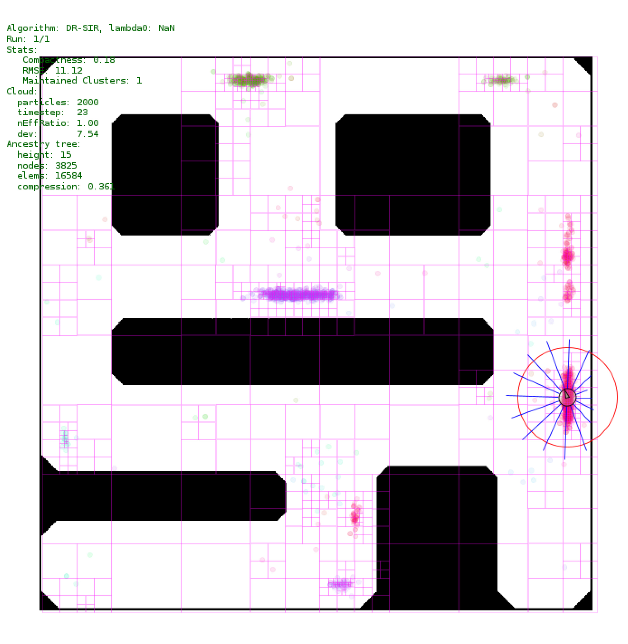}{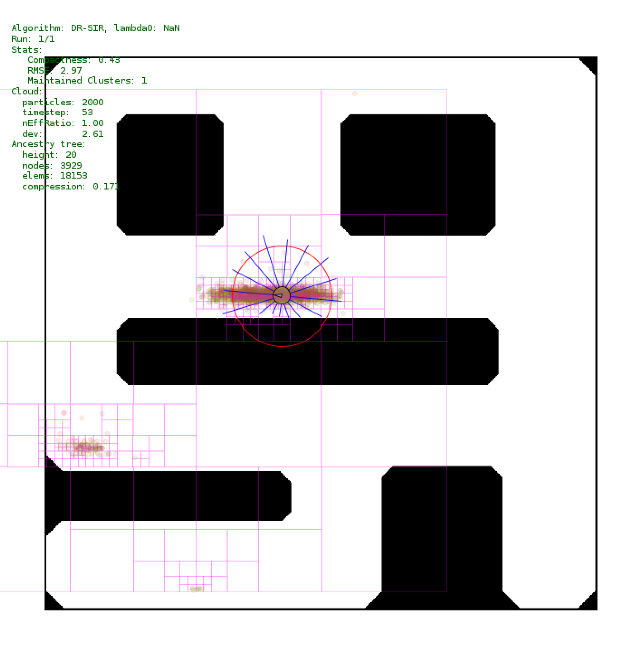}{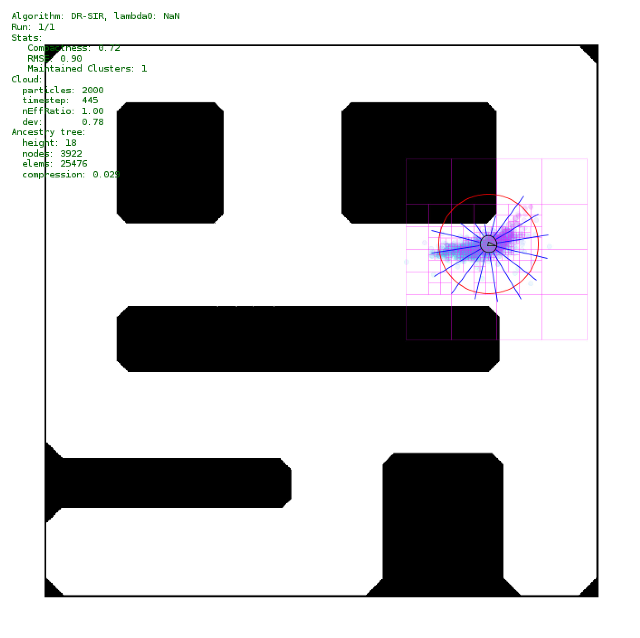}{Deterministic Resampling (DR-SIR) retains multiple modes for a while. After convergence, the estimate stays compact (similar to PF)}

  \label{fig:robotics_visuals_maze_1}
\end{figure*}

\begin{figure*}[!t]
  \centering

  \newcommand{\SubFigImg}[1]{%
      \includegraphics[trim=10 10 10 10, clip, width=0.23\textwidth]{figures/appendix_robot_maze/#1}%
  }

  \newcommand{\FourImageRow}[5]{%
    \begin{minipage}{\textwidth}
        \centering
        \SubFigImg{#1}\hfill
        \SubFigImg{#2}\hfill
        \SubFigImg{#3}\hfill
        \SubFigImg{#4}%
        \caption{#5} % row-level caption (unnumbered)
    \end{minipage}\par\vspace{6pt}
  }

  \FourImageRow{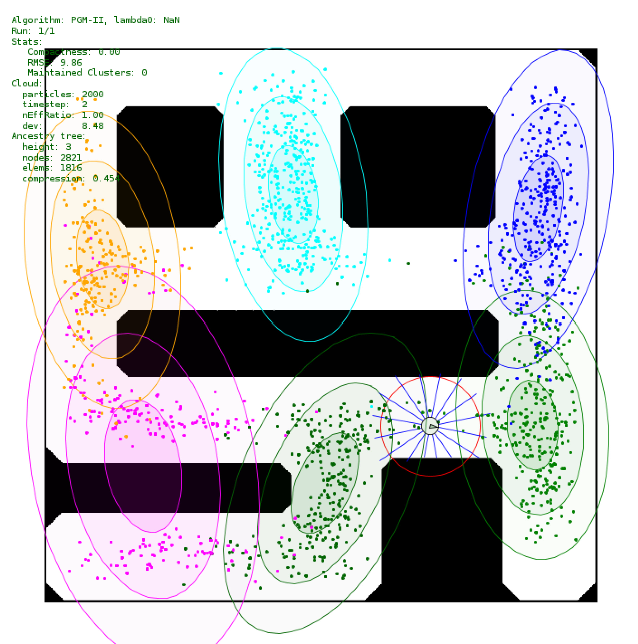}{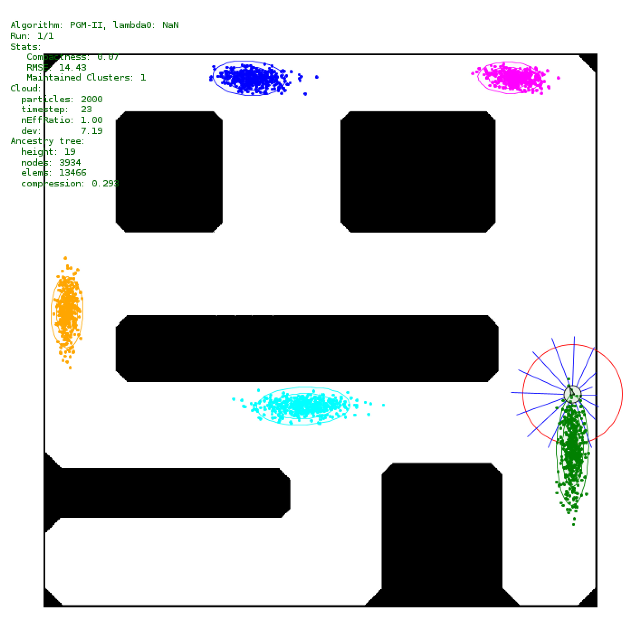}{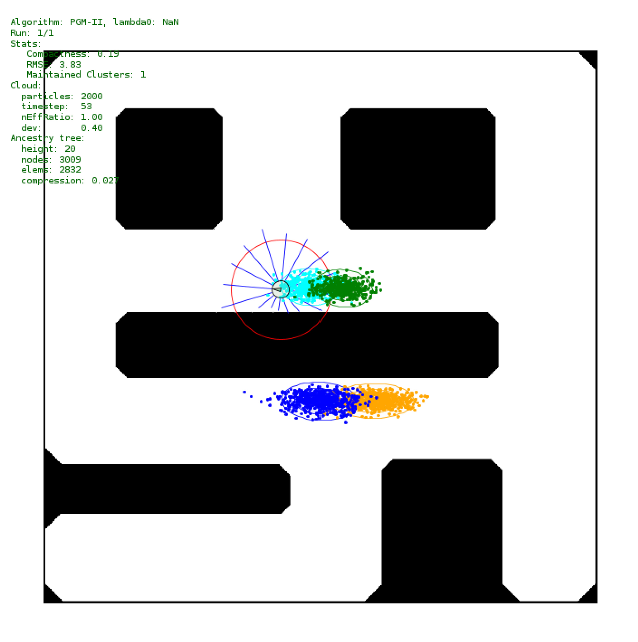}{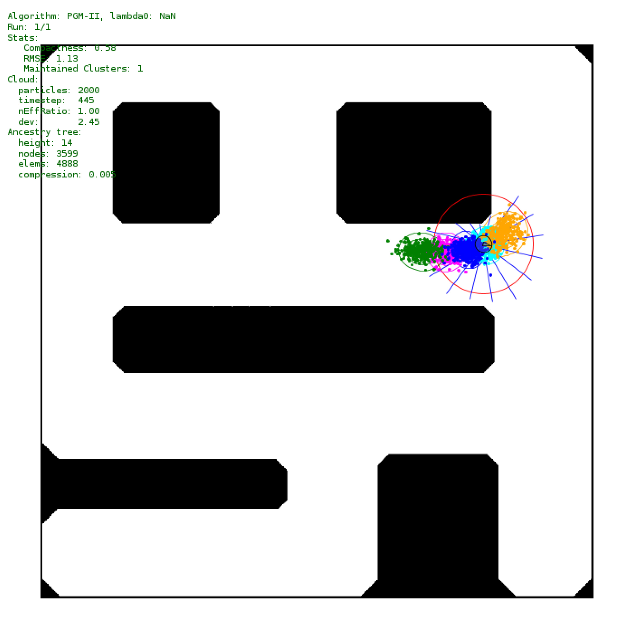}{PGM-II suffers from chaotic initial clustering (left). In this run, it retains multiple modes for a while and manages to converge correctly. Even after convergence, the main mode is split into several clusters. The particle color depicts Gaussian cluster index.}
  \FourImageRow{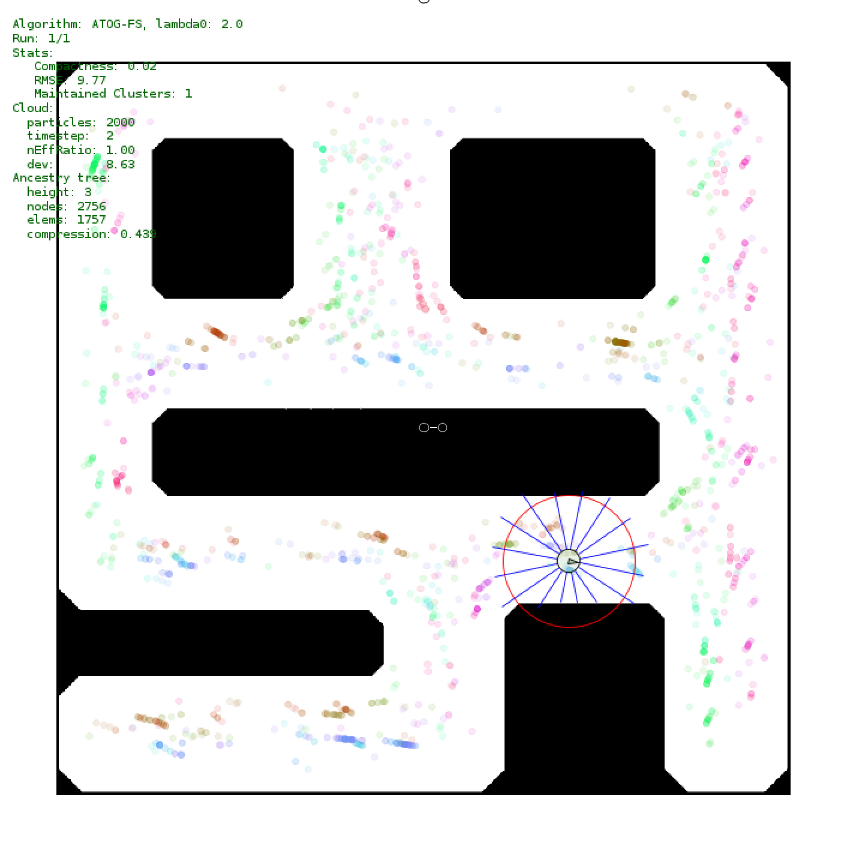}{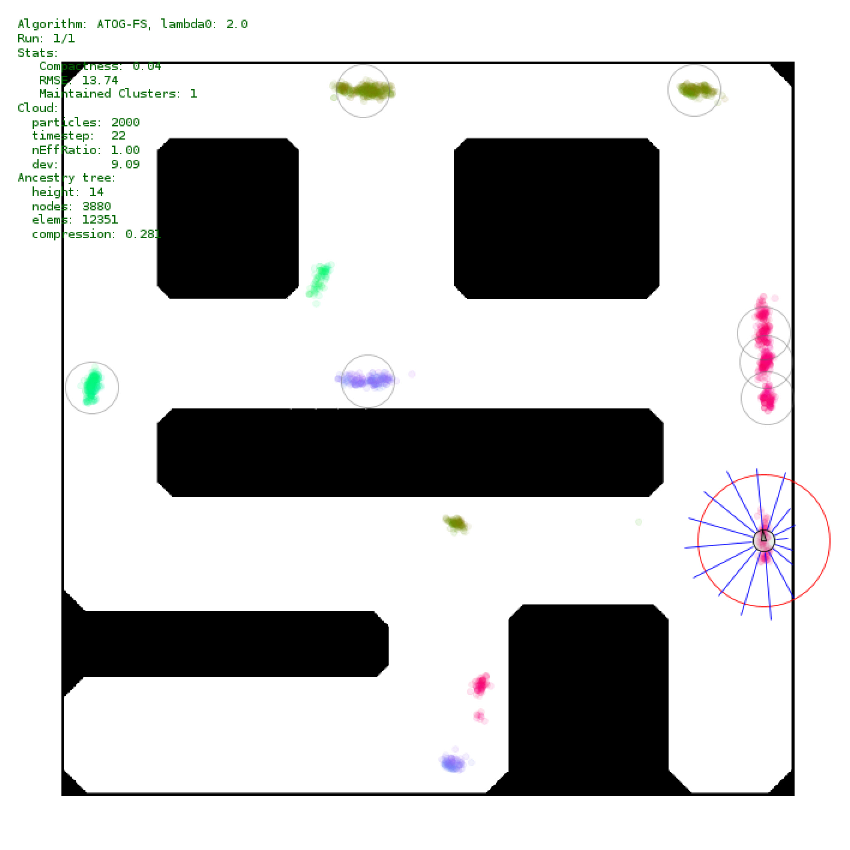}{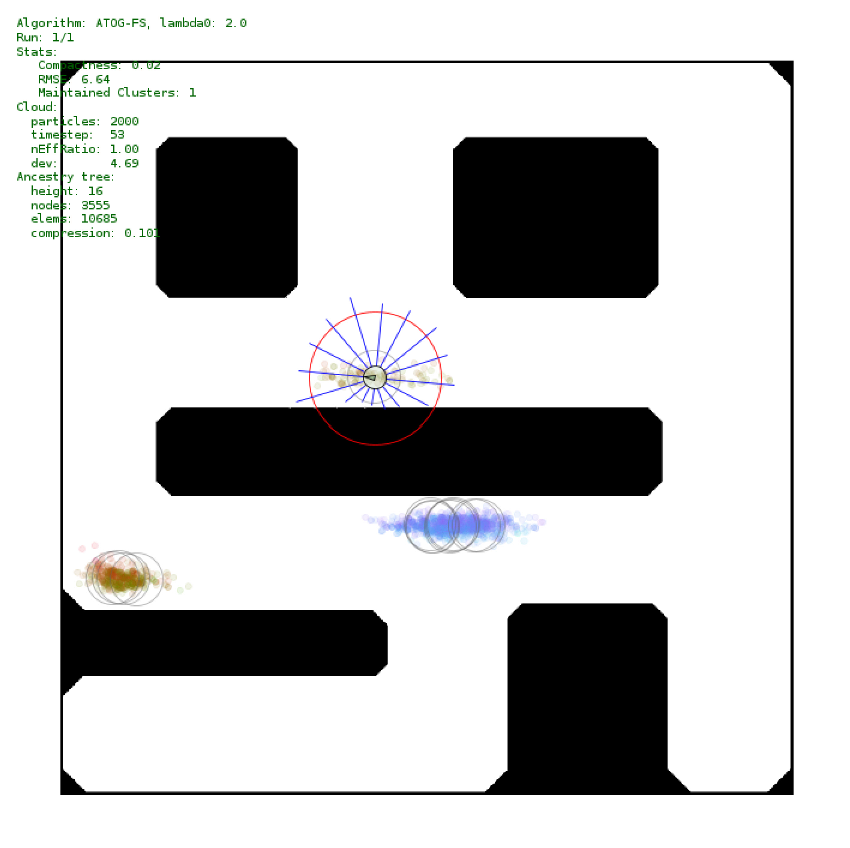}{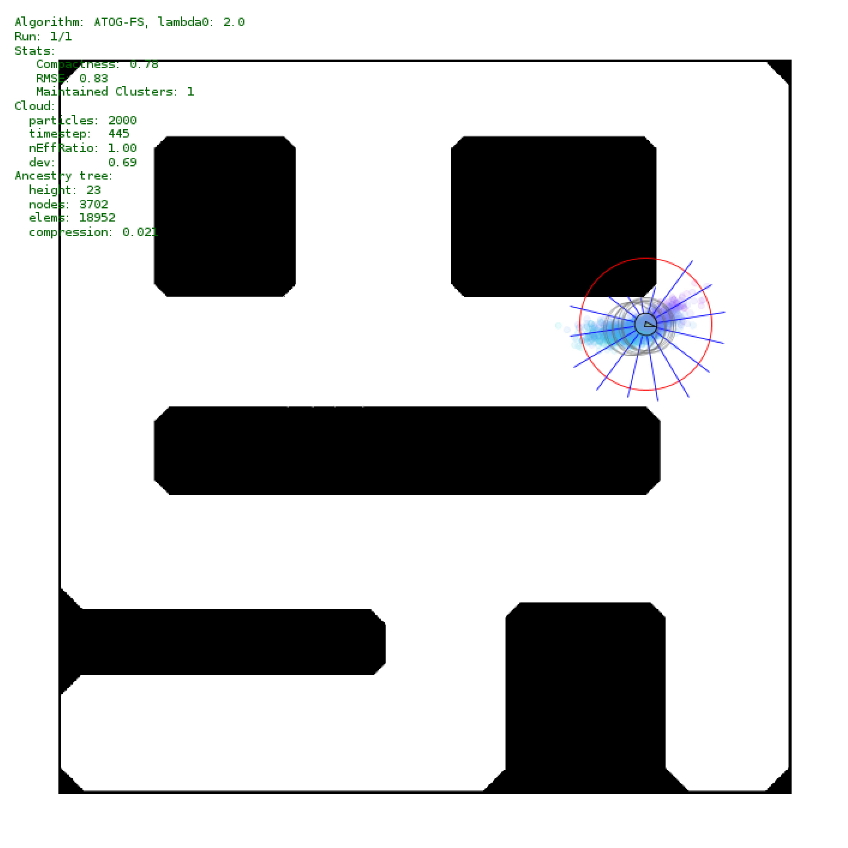}{ATOG-FS retains modes similarly to FDS, but after convergence, keeps the estimate compact (similar to PF).}

  \label{fig:robotics_visuals_maze_2}
\end{figure*}

\section{Ancestry tree metrics over algorithms}
Even when we do not use the ancestry tree for anything (PF and FDS), we can still keep track of it to understand and analyze the convergence behavior of the algorithm. We visualize ancestry tree metrics for selected algorithms to offer preliminary insights into possible future use cases, e.g., detecting multimodality and convergence from the tree topology. Note that for DR-SIR and PGM-II, we cannot form the ancestry tree due to their resampling strategies lacking direct parents. Therefore, those algorithms are omitted from this analysis. We extract metrics from an example run in the Square and Maze environments with $P=2000$ particles. In the Square environment, PF fails to keep track of the four modes, while all other algorithms succeed. In the Maze environment, all algorithms converge correctly. The plots show rolling averages over 100 time steps for readability.

\begin{figure*}[ht]
\centering
\includegraphics[width=0.9\textwidth]{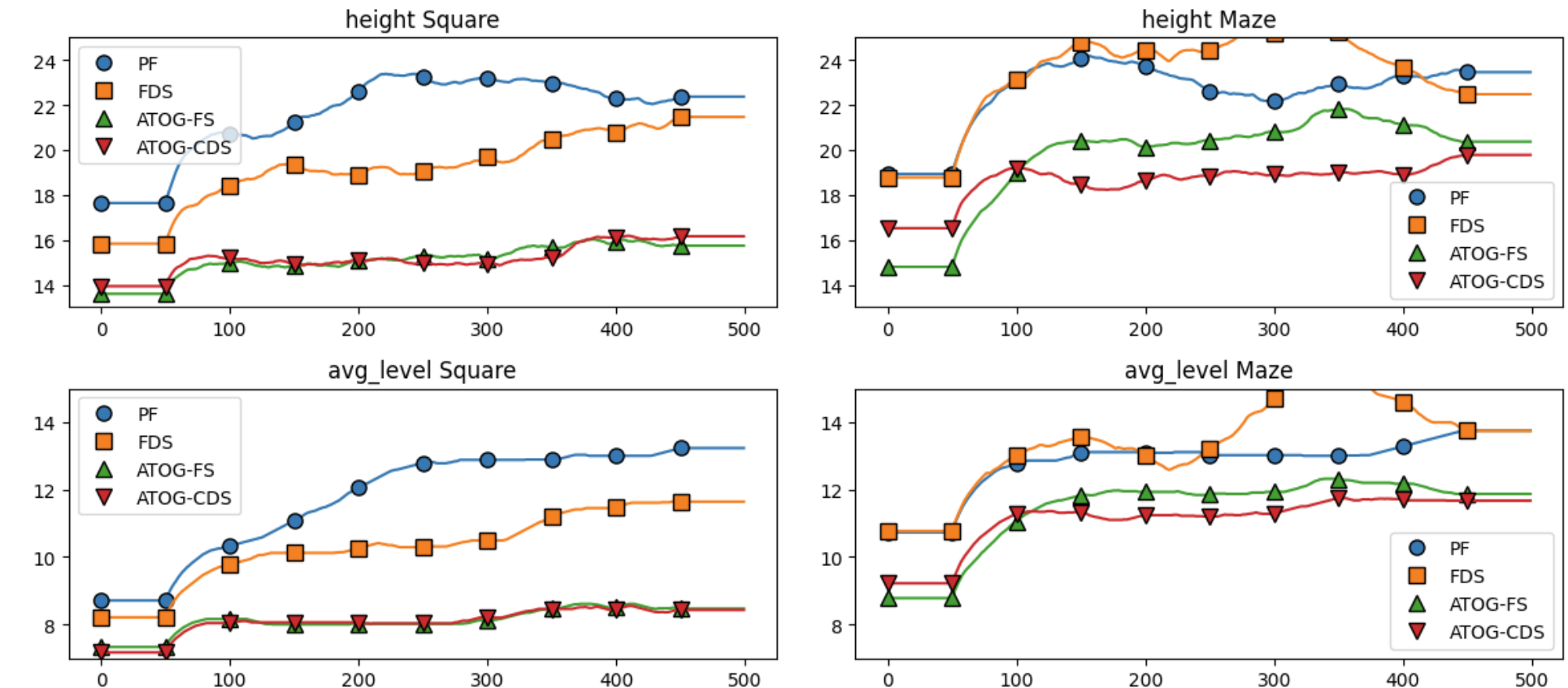}
\caption{Ancestry Tree height metrics from the robotics environments. ATOG variants keep the tree much more shallow.}
\label{fig:appendix_tree_depth}
\end{figure*}

Figure \ref{fig:appendix_tree_depth} visualizes ancestry tree heights and average leaf levels. Especially in the multimodal case, ATOG variants keep the tree much shallower, which we believe corresponds to increased diversity. The standard PF produces the tallest trees. Tree height remains more stable for the ATOG variants.

\begin{figure*}[ht]
\centering
\includegraphics[width=0.9\textwidth]{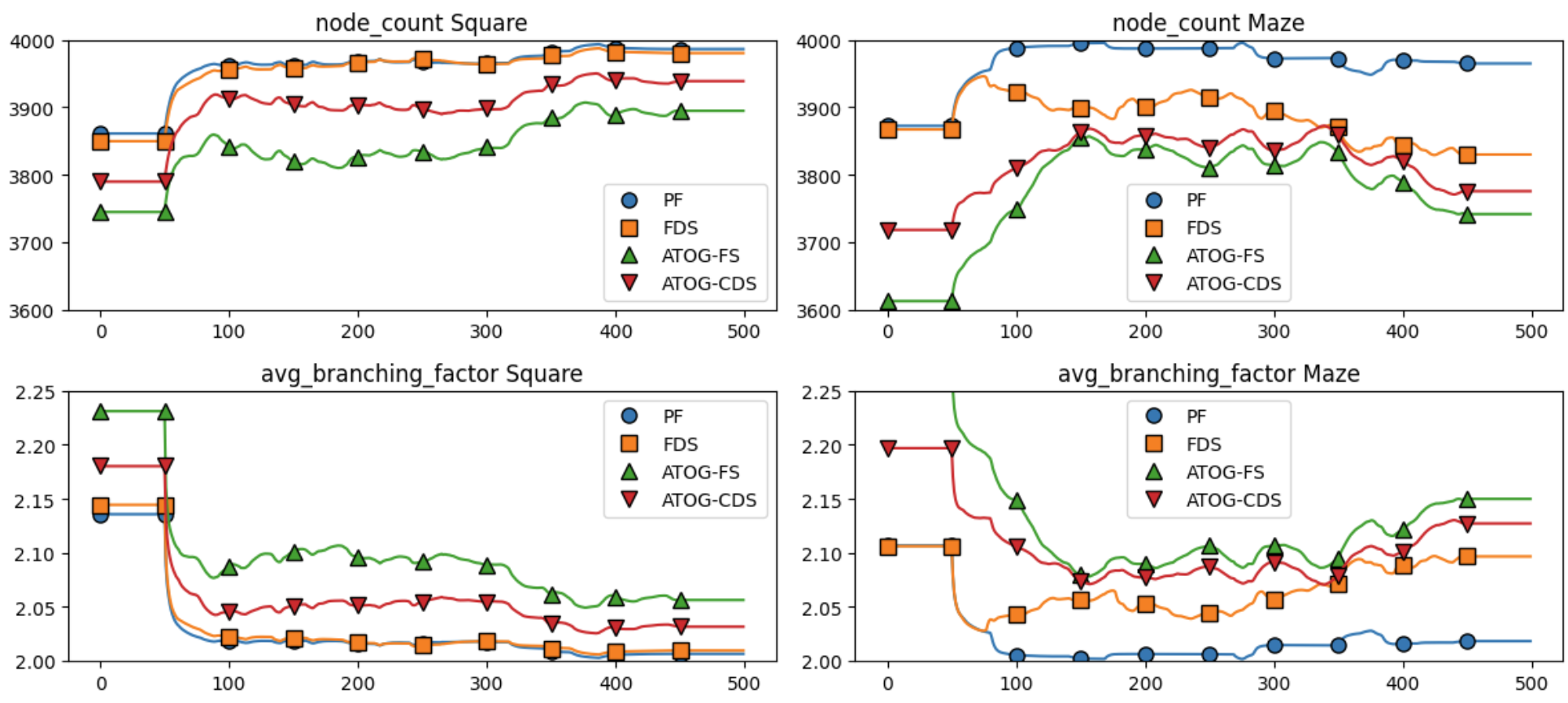}
\caption{Ancestry Tree node metrics from the robotics environments. ATOG variants and FDS clearly react to modality.}
\label{fig:appendix_tree_nodes}
\end{figure*}

Figure \ref{fig:appendix_tree_nodes} depicts tree node counts and branching factors over the run. Since we resample quite aggressively ($N_{\text{eff}} < 0.95P$) and maintain a minimal tree, we expect branching factors close to two. The hard upper bound of $2P$ for tree size is clearly respected. PF appears mostly unaffected by the environment, whereas the other algorithms react to modality.

\begin{figure*}[ht]
\centering
\includegraphics[width=0.9\textwidth]{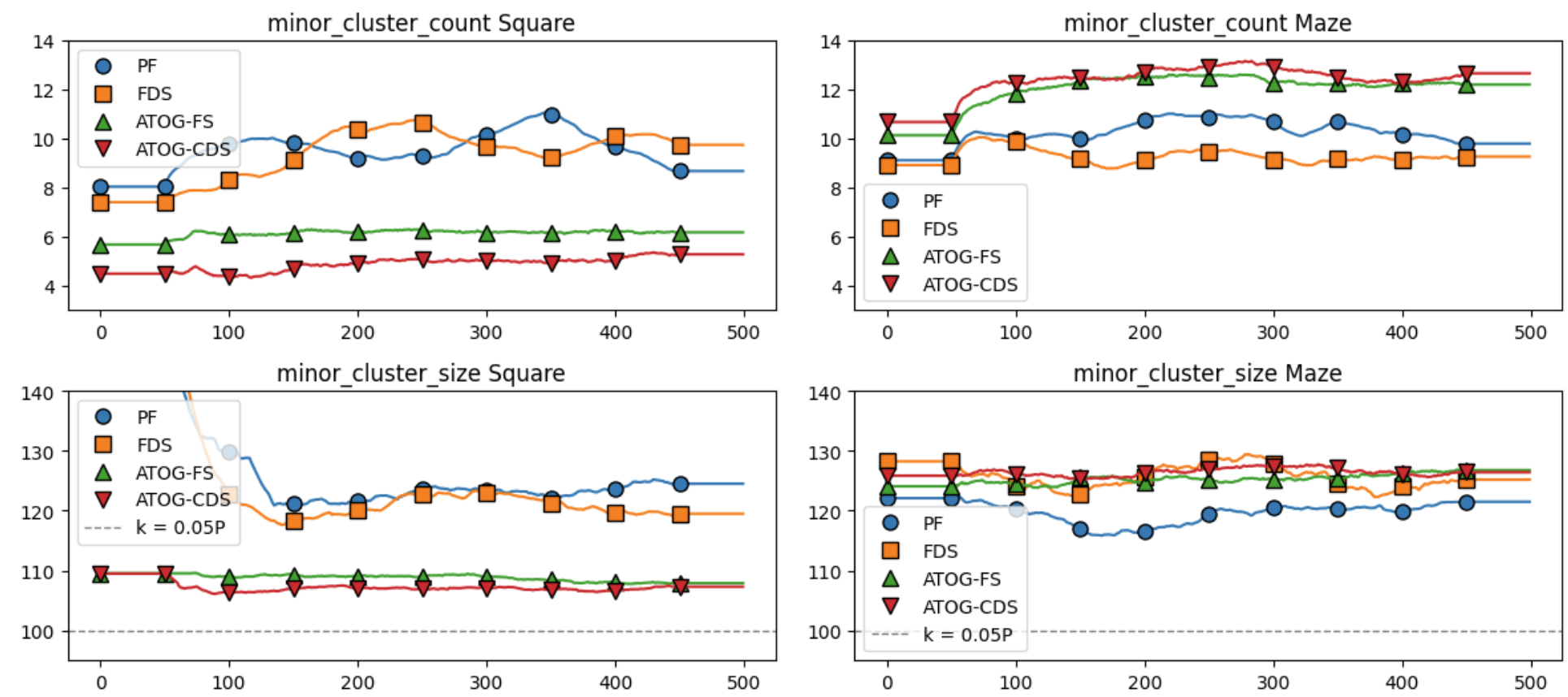}
\caption{Ancestry Tree cluster metrics from the robotics environments. ATOG variant cluster count and size strongly reacts to modality (and convergence). Cluster size consistently falls within $\left[k, 1.5k \right]$.}
\label{fig:appendix_tree_clusters}
\end{figure*}

Figure \ref{fig:appendix_tree_clusters} plots minor cluster counts and average cluster sizes. The cluster sizes remain consistently between $k$ and $1.5k$, and the ATOG variants show more and larger clusters in the unimodal Maze case. This may seem counterintuitive, but it is expected that the converged main mode is represented by multiple small clusters (main branch splits into subtrees).

\end{document}

%% file: commands.tex
\DeclareMathAlphabet{\pazocal}{OMS}{zplm}{m}{n}
\newcommand{\x}{\textbf{x}}
\newcommand{\z}{\textbf{z}}

\renewcommand{\u}{\textbf{u}}

\newcommand{\xIT}{\x_t^{(i)}}
\newcommand{\xITPrev}{\x_{t-1}^{(i)}}

\newcommand{\wIT}{w_t^{(i)}}